\documentclass[runningheads]{llncs}

\usepackage[mobile]{eccv}

\usepackage{eccvabbrv}

\usepackage[accsupp]{axessibility}

\usepackage{graphicx}
\usepackage{booktabs}
\usepackage{makecell}
\usepackage{xcolor}
\usepackage{colortbl}
\usepackage{array}
\usepackage{float}
\usepackage{wrapfig}
\usepackage{enumitem}
\usepackage{algorithm}
\usepackage{algpseudocode}
\usepackage{multirow}
\usepackage[most]{tcolorbox}
\usepackage{afterpage}
\newcommand{\pos}[1]{\\[-1.1ex]{\color{green!60!black}\scriptsize$\uparrow$#1}}
\newcommand{\negdelta}[1]{\\[-1.1ex]{\color{red!70!black}\scriptsize$\downarrow$#1}}

\definecolor{AcademicBlue}{RGB}{45, 85, 133} 
\definecolor{AcademicRed}{RGB}{150, 60, 60}

\newcommand{\dataloss}[1]{\\[-1.1ex]{\color{AcademicBlue}\scriptsize$\downarrow$#1}}
\newcommand{\mecloss}[1]{\\[-1.1ex]{\color{AcademicRed}\scriptsize$\downarrow$#1}}
\newcommand{\gain}[1]{\textbf{\color{green!60!black}#1}}
\newtcolorbox{promptbox}[1]{
    enhanced,
    colback=gray!15,
    colframe=black!80,
    colbacktitle=gray!30,
    coltitle=black,
    fontupper=\fontfamily{ptm}\selectfont\footnotesize,
    fonttitle=\fontfamily{ptm}\selectfont\bfseries,
    title={#1},
    arc=2mm,
    boxrule=1.2pt,
    titlerule=1.2pt,
    left=10pt,
    right=10pt,
    top=8pt,
    bottom=8pt,
    toptitle=3pt,
    bottomtitle=3pt
}

\usepackage[breaklinks,colorlinks,citecolor=eccvblue]{hyperref}

\usepackage{orcidlink}

\begin{document}

\title{OmniJigsaw: Enhancing Omni-Modal Reasoning via Modality-Orchestrated Reordering} 

\titlerunning{OmniJigsaw: Modality-Orchestrated Reordering}

\author{Yiduo Jia\inst{1}$^*$ \and
Muzhi Zhu\inst{1}$^*$ \and
Hao Zhong\inst{1} \and
Mingyu Liu\inst{1} \and
Yuling Xi\inst{1} \and
Hao Chen\inst{1}$^\dag$ \and
Bin Qin\inst{2} \and
Yongjie Yang\inst{2} \and
Zhenbo Luo\inst{2} \and
Chunhua Shen\inst{1}$^\dag$
}

\begingroup
\renewcommand{\thefootnote}{}
\footnotetext{$^*$ Equal contribution. $^\dag$ Corresponding authors.}
\endgroup

\authorrunning{Jia et al.}

\institute{Zhejiang University \and
Xiaomi Inc. \\
{\url{https://aim-uofa.github.io/OmniJigsaw}}
}

\maketitle

\begin{center}
    \captionsetup{type=figure}
    \includegraphics[width=\textwidth]{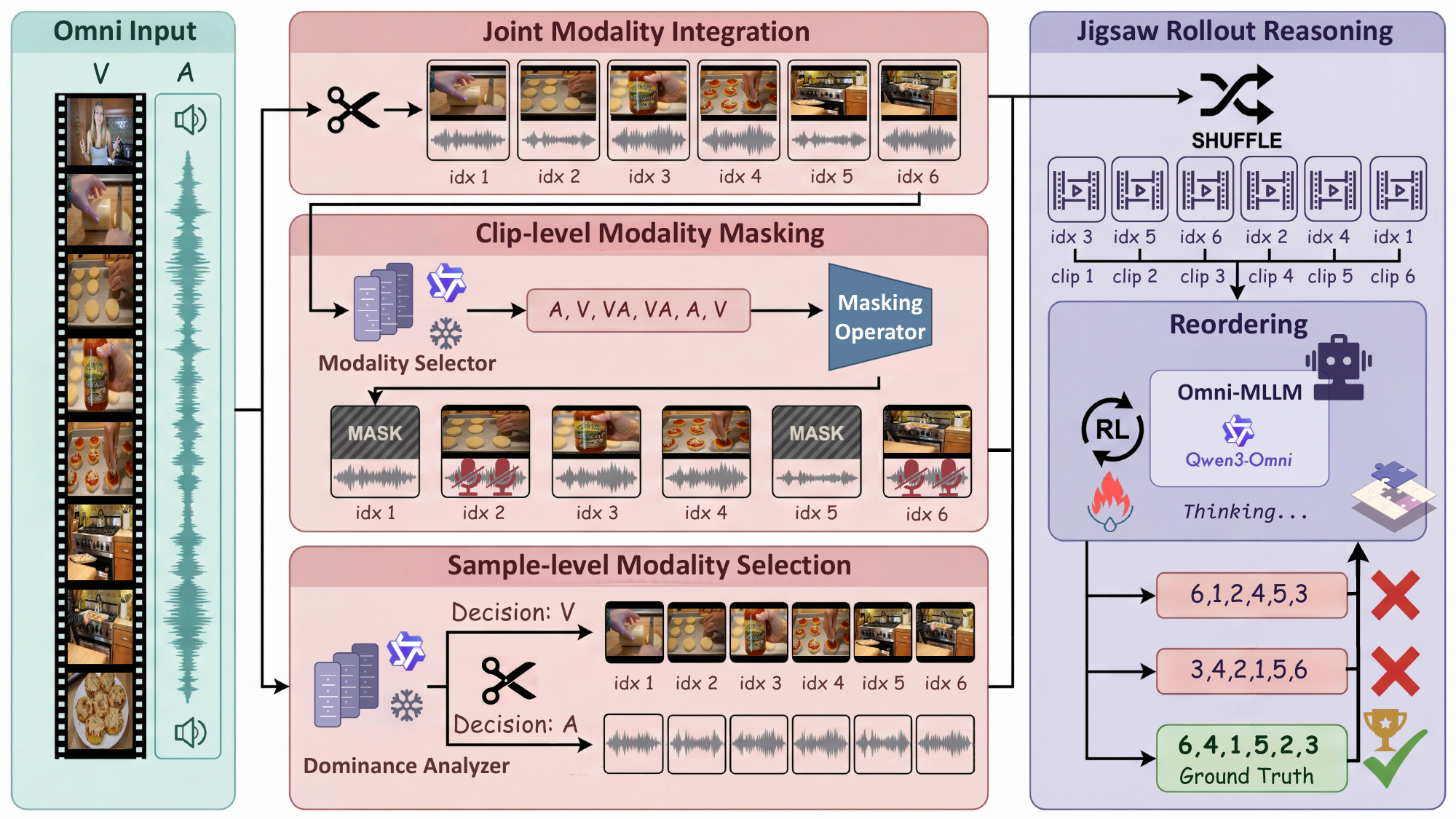}
    \captionof{figure}{\textbf{OmniJigsaw framework for self-supervised omni-modal RL post-training.} Three modality orchestration strategies are illustrated: Joint Modality Integration (JMI, top) leveraging joint audio-visual signals; Clip-level Modality Masking (CMM, center) enforcing an information bottleneck through adaptive clip-wise masking; and Sample-level Modality Selection (SMS, bottom) prioritizing the primary informative modality. Shuffled clips are processed by the model to restore the original sequence order via RL-guided CoT reasoning.}
    \label{fig:omnijigsaw_pipeline}
\end{center}

\begin{abstract}
  To extend the reinforcement learning post-training paradigm to omni-modal models for concurrently bolstering video-audio understanding and collaborative reasoning, we propose OmniJigsaw, a generic self-supervised framework built upon a temporal reordering proxy task. Centered on the chronological reconstruction of shuffled audio-visual clips, this paradigm strategically orchestrates visual and auditory signals to compel cross-modal integration through three distinct strategies: Joint Modality Integration, Sample-level Modality Selection, and Clip-level Modality Masking. Recognizing that the efficacy of such proxy tasks is fundamentally tied to puzzle quality, we design a two-stage coarse-to-fine data filtering pipeline, which facilitates the efficient adaptation of OmniJigsaw to massive unannotated omni-modal data. Our analysis reveals a ``bi-modal shortcut phenomenon'' in joint modality integration and demonstrates that fine-grained clip-level modality masking mitigates this issue while outperforming sample-level modality selection. Extensive evaluations on 15 benchmarks show substantial gains in video, audio, and collaborative reasoning, validating OmniJigsaw as a scalable paradigm for self-supervised omni-modal learning.
    \keywords{Omni-modal Reasoning \and Reinforcement Learning \and MLLM}
\end{abstract}

\section{Introduction}
\label{sec:intro}
The ultimate goal of Artificial General Intelligence (AGI) is to develop intelligent agents capable of comprehensively processing omni-modal inputs, spanning video, audio, and text, to perform complex reasoning, decision-making, and planning~\cite{li2025comprehensive,yue2025simulating}. Recently, reinforcement learning (RL) post-training~\cite{ouyang2022training,rafailov2023direct,lambert2024tulu} has driven remarkable breakthroughs in large language models (LLMs), empowering them with robust reasoning capabilities to solve intricate mathematical problems~\cite{grattafiori2024llama,yang2024qwen2,guo2025deepseek} and generate high-quality, functional code~\cite{roziere2023code,zhu2024deepseek}. Despite these significant advancements in purely textual domains, how to thoroughly explore and effectively enhance the reasoning capabilities of omni-modal models~\cite{xu2025qwen25omnitechnicalreport,xu2025qwen3} within the context of integrated multi-modal processing remains an open and challenging problem~\cite{chen2026omnivideo,zhou2025reinforced,li2025perception}.

A primary bottleneck impeding the extension of these RL-driven successes to omni-modal reasoning is the significant difficulty of acquiring massive, high-quality annotated data and providing effective supervisory signals. In text-only domains such as mathematics or coding, it is relatively straightforward to generate large-scale problem instances and provide verifiable, deterministic feedback for RL optimization~\cite{wang2025reinforcement}. Conversely, for omni-modal understanding~\cite{wang2025test,tu2025favor,ccoban2024mllms,guan2025mllm}, collecting an equivalent volume of omni-modal data that intrinsically necessitates complex collaborative cross-modal reasoning is prohibitively expensive and labor-intensive~\cite{zhang2025video,wang2025cotasks,jiang2025videop2r}. Driven by these challenges, we explore a fundamental question in this work: Can we identify a suitable proxy task that effectively leverages massive unannotated omni-modal data to bolster the versatile reasoning capabilities of omni-modal models via a self-supervised paradigm?

Inspired by the efficacy of jigsaw-based RL post-training in the visual domain~\cite{wu2025visual}, we pioneer the extension of this paradigm into the generalized audio-visual domain, investigating whether an omni-modal model can enhance its reasoning capabilities through the chronological reordering of shuffled clips. Initially, we design a straightforward Joint Modality Integration (JMI) strategy that provides full accessibility to both visual and auditory streams. However, we counter-intuitively observe a ``bi-modal shortcut phenomenon'' where modality-specific cues independently suffice to solve the task, triggering a modal short-circuit that allows the model to take the path of lower resistance, thereby hindering the robust cultivation of reasoning for the weaker modality. To address this, we further propose two ingenious orchestration strategies: (1) Sample-level Modality Selection (SMS), which deploys the model as a global dominance analyzer to identify the primary modality and mitigate interference from less informative streams; and (2) Clip-level Modality Masking (CMM), which utilizes the model as a dynamic modality selector to evaluate the semantic density within each clip and selectively mask the less salient modality, thereby intentionally constructing a cross-modal information bottleneck that forces the model to integrate fragmented heterogeneous signals to reconstruct the global timeline.

Additionally, given that the efficacy of such jigsaw-based proxy tasks is fundamentally predicated on the solvability and quality of the generated puzzles, we establish a two-stage coarse-to-fine data filtering pipeline combining signal-level heuristic filtering with MLLM-based semantic Chain-of-Thought (CoT) screening, which guarantees the temporal irreversibility and clear state transitions inherently required by the jigsaw task. By transforming the resource-intensive annotation process typically reliant on heavy teacher models~\cite{zhang2025video,wang2025cotasks,jiang2025videop2r} into a lightweight data-filtering workflow, we provide robust support for the efficient adaptation of OmniJigsaw to massive unannotated omni-modal data. Concurrently, we design a composite reward mechanism comprising positional and adjacency accuracy metrics alongside format rewards and repetition penalties, while further introducing an accuracy-dependent discount factor that effectively suppresses sub-optimal solutions and catalyzes the model to actively pursue perfect chronological restoration.

Extensive experiments demonstrate that OmniJigsaw yields substantial enhancements across uni-modal video reasoning, audio comprehension, and collaborative omni-modal reasoning. Notably, our CMM strategy boosts the robust Qwen3-Omni-30B-A3B-Instruct~\cite{xu2025qwen3} baseline by significant margins, achieving absolute gains of \textbf{+4.38} on MLVU-Test~\cite{zhou2025mlvu}, \textbf{+2.50} on MMAR~\cite{ma2025mmar}, and \textbf{+1.70} on OmniVideoBench~\cite{li2025omnivideobench}. Further rich ablations and analyses compellingly validate the efficacy of our meticulously designed data pipeline and reward mechanisms.

In summary, our main contributions are:
\begin{enumerate}[label=\textbf{\arabic*)}]
    \item We pioneer the extension of jigsaw-based RL post-training to omni-modal reasoning by proposing OmniJigsaw, a generic, lightweight, and annotation-free self-supervised framework that leverages modality-orchestrated temporal reordering to bolster complex reasoning capabilities.
    \item We identify and deeply analyze the ``bi-modal shortcut phenomenon'' inherent in the Joint Modality Integration (JMI) strategy, and consequently propose two advanced orchestration strategies: Sample-level Modality Selection (SMS) and Clip-level Modality Masking (CMM), which further enhance the reasoning performance of omni-modal models.
    \item We establish a two-stage coarse-to-fine data filtering pipeline that facilitates the efficient adaptation of our framework to massive unannotated omni-modal data, thereby significantly enhancing its scalability.
    \item We conduct comprehensive ablation studies and analyses demonstrating the sensitivity of omni-modal jigsaw proxy tasks to data quality, reward mechanisms, and the granularity of modality orchestration, thereby offering valuable empirical insights for future research in self-supervised omni-modal learning.
\end{enumerate}

\section{Related Work}
\label{sec:rel_work}

\subsection{RL Post-Training for Omni-Modal Understanding}
While RL post-training has evolved from prioritizing human intent alignment (\eg, RLHF~\cite{ouyang2022training}, DPO~\cite{rafailov2023direct}) to comprehensively strengthening complex reasoning (\eg, RLVR~\cite{lambert2024tulu}) across textual~\cite{lambert2024tulu,grattafiori2024llama,yang2024qwen2,guo2025deepseek,tunstall2023zephyr,roziere2023code,zhu2024deepseek} and visual domains (\eg, VideoChat-R1~\cite{li2025videochat}, RLHF-V~\cite{yu2024rlhf}, Visual-RFT~\cite{liu2025visual}, Diffusion-DPO~\cite{wallace2024diffusion}), its potential to simultaneously enhance video and audio reasoning capabilities in omni-modal models remains insufficiently explored~\cite{chen2026omnivideo,zhou2025reinforced,li2025perception}. Despite the rapid advancement of specialized video~\cite{wang2025test,tu2025favor,yang2026mllm} and audio~\cite{ccoban2024mllms,yang2025omni,guan2025mllm} understanding alongside integrated architectures like Qwen-Omni~\cite{xu2025qwen25omnitechnicalreport,xu2025qwen3} driven by the surging demand for holistic perception in embodied AI~\cite{li2025comprehensive,yue2025simulating}, current audio-visual enhancements predominantly rely on computationally intensive supervised training with meticulously annotated data (\eg, Video-CoT~\cite{zhang2025video}, CoTasks~\cite{wang2025cotasks}, VIDEOP2R~\cite{jiang2025videop2r}), complex auxiliary objectives leveraging external reward models~\cite{zhao2025unified} (\eg, VideoWorld 2~\cite{ren2026videoworld2learningtransferable}, Dual-IPO~\cite{yang2026dualipodualiterativepreferenceoptimization}), or elaborate multi-stage RL pipelines like Omni-R1~\cite{zhong2025omni}. To bridge this gap without necessitating costly manual annotation or architectural complexity, our OmniJigsaw framework introduces a lightweight and verifiable self-supervised proxy task that strategically orchestrates synchronized video and audio streams to concurrently bolster intrinsic omni-modal comprehension and collaborative reasoning capabilities.

\subsection{Jigsaw as Self-Supervised Proxy Task}
With the evolution of self-supervised learning, identifying effective proxy tasks that distill supervisory signals directly from the innate data topology without manual annotation remains a central challenge, bringing prominence to jigsaw-style tasks characterized by concise objectives, computational efficiency, and independence from auxiliary generative models. Pioneered by Noroozi and Favaro~\cite{noroozi2016unsupervised} to compel the learning of object parts and spatial layouts in the static visual domain, this permutation-based philosophy has transcended diverse modalities, extending to temporal order verification for capturing motion dynamics in videos~\cite{misra2016shuffle}, rearranged voxel reconstruction for bolstering spatial reasoning on 3D point clouds~\cite{sauder2019self}, permutation language modeling for bidirectional context dependencies in NLP~\cite{yang2019xlnet}, and multimodal puzzles for robust cross-modal alignment in medical imaging~\cite{taleb2021multimodal}. While recent efforts have demonstrated the efficacy of visual-only jigsaw tasks~\cite{wu2025visual}, OmniJigsaw surpasses these uni-modal boundaries by introducing fine-grained modality orchestration strategies, designed to further unlock the potential of structural ordering tasks within the generalized video-audio domain.

\section{Method}
\label{sec:method}

\subsection{OmniJigsaw Formulation}
\label{sec:omnijigsaw_formulation}
We formally define the omni-modal jigsaw task as a permutation prediction problem over an omni-modal temporal sequence. Let $\mathcal{X} = (\mathcal{V}, \mathcal{A})$ denote a video sample comprising a visual stream $\mathcal{V} \in \mathbb{R}^{T \times H \times W \times 3}$ and a raw audio waveform $\mathcal{A} \in \mathbb{R}^{L}$. We first segment $\mathcal{X}$ uniformly along the temporal dimension into $N$ non-overlapping clips. To prevent trivial solutions derived from low-level boundary continuity, we apply a trimming operation $\mathcal{T}_{trim}$ to discard a temporal span from both the beginning and the end of each clip, involving both frames and audio signals. This yields an ordered sequence of discrete omni-modal segments $\mathcal{S} = [s_1, s_2, \dots, s_N]$, where $s_i = (v_i, a_i)$ represents the $i$-th clip containing synchronized visual and acoustic information. Let $\pi$ be a random permutation sampled from the set of bijective mappings $\{1, \dots, N\} \to \{1, \dots, N\}$. The disordered input sequence $\tilde{\mathcal{S}}$ is constructed as $\tilde{\mathcal{S}} = [\tilde{s}_1, \tilde{s}_2, \dots, \tilde{s}_N]$, where $\tilde{s}_j = s_{\pi^{-1}(j)}$ for all $j \in \{1, \dots, N\}$. The objective of the omni-modal model $\mathcal{M}_\theta$ is to restore the original chronological order given the shuffled inputs. Specifically, the model is tasked to predict a sequence of indices $\hat{\mathbf{y}}$ that aligns with the ground truth permutation sequence $\mathbf{y}$, which is $[\pi(1), \pi(2), \dots, \pi(N)]$. The generalized optimization objective is formulated as:
\begin{equation}
    \hat{\mathbf{y}} = \mathcal{M}_\theta(\Phi(\tilde{\mathcal{S}}); \mathcal{I}_{prompt}), \quad \text{s.t.} \quad \hat{\mathbf{y}} \rightarrow \mathbf{y},
\end{equation}
where $\mathcal{I}_{prompt}$ denotes the task instruction, and $\Phi(\cdot)$ acts as a strategy-specific orchestration function that governs the modality accessibility and masking mechanism for each clip within the sequence. The model is required to explicitly output the chain of thought followed by the predicted indices. Figure~\ref{fig:omnijigsaw_pipeline} illustrates the overall pipeline of our proposed OmniJigsaw framework and three modality orchestration strategies.

\subsection{Joint Modality Integration Strategy}
\label{sec:jmi}
Trivially preserving the integrity of omni-modal information, Joint Modality Integration (JMI) strategy compels the omni-modal model to concurrently analyze visual scene evolution and auditory cues within each clip to reconstruct the global timeline. In this setting, the strategy-specific orchestration function $\Phi_{jmi}$ acts as an identity mapping that retains the complete synchronized visual and acoustic information for all clips in the disordered sequence. Specifically, we define a temporal downsampling operator $\mathcal{D}_T$ applied to the visual stream to obtain a sparse representation while maintaining synchronization with the audio track. The jigsaw rollout process under this strategy is thus specialized as $\mathcal{M}_\theta(\Phi_{jmi}(\tilde{\mathcal{S}}); \mathcal{I}_{jmi})$, where
\begin{equation}
    \Phi_{jmi}(\tilde{s}_j) = \mathcal{D}_T(\tilde{s}_j) = (\mathcal{D}_T(v_{\pi^{-1}(j)}), a_{\pi^{-1}(j)}),
\end{equation}
which necessitates the model to leverage correlative evidence from both modalities to predict the target permutation $[\pi(1), \dots, \pi(N)]$.

\subsection{Sample-level Modality Selection Strategy}
\label{sec:sms}
Recognizing that some videos are inherently visually dominant while others are acoustically driven, Sample-level Modality Selection (SMS) strategy operates at the sample level through a global decision mechanism to select the appropriate modality for the omni-jigsaw task. Specifically, this strategy first employs the model as a dominance analyzer to process the complete raw audio stream $\mathcal{A}$ alongside the globally downsampled visual stream $\mathcal{D}_T(\mathcal{V})$ to determine the primary information and temporal carrier $d \in \{V, A\}$ via the decision function:
\begin{equation}
    d = \operatorname*{arg\,max}_{m \in \{V, A\}} p_\theta(m \mid (\mathcal{D}_T(\mathcal{V}), \mathcal{A}); \mathcal{I}_{judge}). 
\end{equation}
Subsequently, the jigsaw rollout process is instantiated exclusively on the selected modality to mitigate interference from the less informative stream, formulated as $\mathcal{M}_\theta(\Phi_{sms}(\tilde{\mathcal{S}} | d); \mathcal{I}_{sms})$, where the orchestration function $\Phi_{sms}$ applies a modality-conditional projection defined as:
\begin{equation}
    \Phi_{sms}(\tilde{s}_j | d) = 
    \begin{cases} 
    (\mathcal{D}_T(v_{\pi^{-1}(j)}), \mathbf{0}), & \text{if } d = V \\
    (\mathbf{0}, a_{\pi^{-1}(j)}), & \text{if } d = A 
    \end{cases},
\end{equation}
which ensures that the reordering reasoning is conducted solely on the modality containing the most salient temporal signals, thereby avoiding noise from the modality characterized by sparse information or lacking irreversible temporal flow.

\begin{figure}[t]
    \centering
    \includegraphics[width=\linewidth]{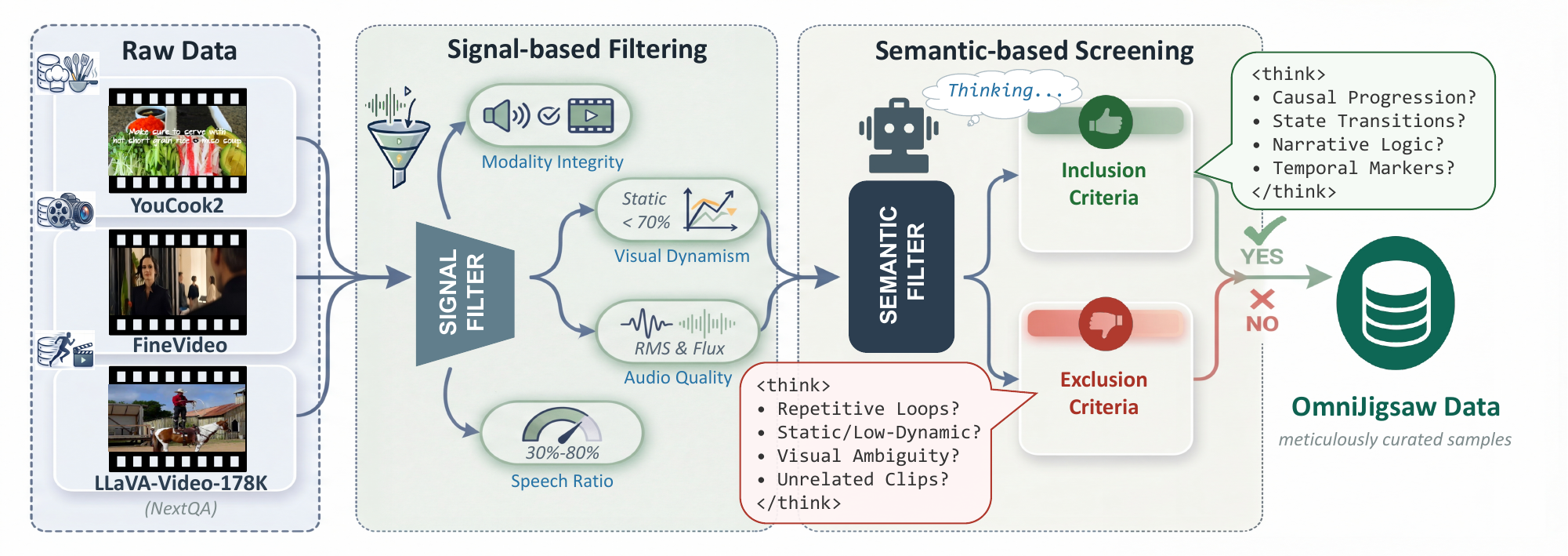}
    \caption{\textbf{Data filtering pipeline for efficient adaptation of OmniJigsaw.} Raw videos are subjected to signal-based filtering to ensure omni-modal integrity and dynamism, followed by semantic-based screening incorporating CoT reasoning for the assessment of narrative logic and state transitions.}
    \label{fig:data_pipeline}
\end{figure}

\subsection{Clip-level Modality Masking Strategy}
\label{sec:cmm}
Clip-level Modality Masking (CMM) strategy introduces an information bottleneck by selectively masking modalities based on their semantic density to foster robust cross-modal reasoning capabilities through a two-stage process. In the first stage, the model functions as a modality selector that evaluates the information richness and temporal criticality of each clip within the ordered sequence $\mathcal{S}$ to generate a modality selection vector $\mathbf{m} = [m_1, \dots, m_N]$, where $m_i \in \{V, A, VA\}$ denotes retaining only video, retaining only audio, or retaining both modalities, respectively. In the second stage, we apply a masking operator $\mathcal{M}_{cmm}$ parameterized by the selection decision that retains the original feature tensor if the modality is chosen by the selector, while replacing the unselected modality with a null tensor $\mathbf{0}$. The orchestration function $\Phi_{cmm}$ applies this operator to each disordered clip $\tilde{s}_j$ based on its corresponding decision $m_{\pi^{-1}(j)}$. The jigsaw rollout process is formally expressed as $\mathcal{M}_\theta(\Phi_{cmm}(\tilde{\mathcal{S}} | \mathbf{m}); \mathcal{I}_{cmm})$, where
\begin{equation}
    \Phi_{cmm}(\tilde{s}_j | m_{\pi^{-1}(j)}) = \mathcal{M}_{cmm}(\mathcal{D}_T(\tilde{s}_j) | m_{\pi^{-1}(j)}).
\end{equation}
This mechanism imposes an information bottleneck that forces the model to switch its attention dynamically between visual and acoustic clues to recover the target permutation sequence.

\subsection{Data Filtering Pipeline}
\label{sec:data_pipeline}

To ensure puzzle solvability, we design a two-stage data filtering pipeline to exclude ill-posed samples lacking modal integrity or irreversible transitions that provide insufficient temporal reordering cues.

\subsubsection{Signal-based Heuristic Filtering}
\label{sec:data_s1}
Initially, we employ heuristic lightweight algorithms to efficiently prune low-quality samples, the workflow of which is depicted in the first stage of Fig.~\ref{fig:data_pipeline}. Prioritizing omni-modal integrity, we discard instances missing either visual or audio streams. To guarantee visual dynamism, we calculate the Mean Absolute Difference (MAD) between adjacent frames and filter out videos where static scenes exceed $70\%$. Regarding audio quality, we utilize Root Mean Square (RMS) amplitude and Spectral Flux (SF) variance to remove silence or constant noise. Furthermore, a Silero Voice Activity Detection (VAD) model is applied to ensure the speech ratio lies between $30\%$ and $80\%$, balancing information density with visual diversity.

\subsubsection{Semantic-based Reasoning Screening}
\label{sec:data_s2}
Building upon the signal-based filtering, we utilize a lightweight MLLM for semantic assessment, transforming resource-intensive annotation by heavy teacher models into an efficient filtering workflow. As illustrated in Fig.~\ref{fig:data_pipeline} (stage 2), the model is prompted to identify irreversible temporal flows and clear state transitions, while excluding repetitive loops, low-dynamic content, or visually ambiguous scenes. To enhance reliability, a prompt-based CoT mechanism requires the model to articulate its logic within \texttt{<think>} tags before outputting a final \texttt{YES/NO} decision. Only samples validated with a coherent CoT and a \texttt{YES} decision are retained to ensure the curated data exhibit strong temporal-causal structures suitable for the jigsaw task.

\subsection{Reward Design}
\label{sec:reward_design}
To effectively guide the policy optimization under the OmniJigsaw framework, we construct a composite reward function $R_{tot}$ to steer the model towards precise reordering and structural adherence:
\begin{equation}
    R_{tot} = R_{rep} + R_{fmt} + \lambda(\text{acc}) \cdot \left( w_{pos} \cdot R_{pos} + w_{cont} \cdot R_{cont} \right).
\end{equation}
To mitigate repetitive loops during reasoning, a penalty $R_{rep}$ of $-0.5$ is applied based on excessive N-gram ($N=20$) repetitions exceeding a threshold of $3$. Simultaneously, to incentivize deep deliberation and ensure parsing reliability, a format reward $R_{fmt}$ of $+0.2$ is awarded for strict adherence to \texttt{<think>...</think><answer>...</answer>}. Regarding the reordering precision, let $\hat{\mathbf{y}} = [\hat{y}_1, \dots, \hat{y}_N]$ denote the predicted index sequence. We define the global positional accuracy $R_{pos}$ to measure the absolute correctness of clip placement, and the local continuity accuracy $R_{cont}$ to reward partial reasoning success and cross-modal alignment through the preservation of adjacent pairs:
\begin{equation}
    R_{pos} = \frac{1}{N} \sum_{i=1}^N \mathbb{I}(\hat{y}_i = y_i), \quad R_{cont} = \frac{1}{N-1} \sum_{i=1}^{N-1} \mathbb{I}\left((\hat{y}_i, \hat{y}_{i+1}) = (y_i, y_{i+1})\right).
\end{equation}
Furthermore, to encourage the model to pursue the perfect solution $\mathbf{y}$ rather than settling for sub-optimal local minima, we introduce an accuracy-dependent discount factor $\lambda(\text{acc})$, which is set to $1.0$ for a perfect match ($\hat{\mathbf{y}} = \mathbf{y}$) and discounted to $0.2$ otherwise.

\section{Experiments}
\label{sec:experiments}

\begin{table}[!t]
    \centering
    \caption{\textbf{Performance comparison on video reasoning.} Results are reported across two inference modes ($w/$ and $w/o$ audio). \textbf{Bold}: best; {\color{green!60!black}green}: gains over baseline.}
    \label{tab:video_table}

    \fontfamily{ptm}\selectfont
    
    \renewcommand{\arraystretch}{1.2}
    \setlength{\tabcolsep}{3.5pt}
    
    \footnotesize
    \begin{tabular}{@{} l *{8}{c} @{}}
        \toprule[1.2pt]
        \textbf{Method} &
        \rotatebox{60}{\shortstack[c]{\textbf{\scriptsize AoT}\\\textbf{\scriptsize Bench}}} & 
        \rotatebox{60}{\shortstack[c]{\textbf{\scriptsize TUNA}\\\textbf{\scriptsize -Bench}}} & 
        \rotatebox{60}{\shortstack[c]{\textbf{\scriptsize Temp}\\\textbf{\scriptsize Compass}}} & 
        \rotatebox{60}{\shortstack[c]{\textbf{\scriptsize Video}\\\textbf{\scriptsize -TT}}} & 
        \rotatebox{60}{\shortstack[c]{\textbf{\scriptsize Video}\\\textbf{\scriptsize -Holmes}}} & 
        \rotatebox{60}{\shortstack[c]{\textbf{\scriptsize MLVU}\\\textbf{\scriptsize -Test}}} & 
        \rotatebox{60}{\shortstack[c]{\textbf{\scriptsize Video}\\\textbf{\scriptsize -MME}}} & 
        \rotatebox{60}{\textbf{\scriptsize MLVU}} \\
        \midrule
        \multicolumn{9}{c}{\textit{Infer w/ Audio}} \\
        Omni-R1         & 52.09 & 53.84 & 63.10 & 36.80 & 40.72 & 52.59 & 63.20 & 65.69 \\
        HumanOmniV2     & 48.58 & 49.16 & 63.86 & 40.30 & 42.90 & 49.00 & 65.60 & 66.70 \\
        \rowcolor{gray!15} Qwen3-Omni-30B & 64.88 & 62.57 & 70.63 & 44.30 & 50.84 & 57.97 & 72.10 & 70.01 \\
        \midrule
        VideoJigsaw     & \makecell{67.45 \pos{2.57}} & \makecell{63.41 \pos{0.84}} & \makecell{\textbf{72.28} \pos{1.65}} & \makecell{44.90 \pos{0.60}} & \makecell{51.99 \pos{1.15}} & \makecell{60.16 \pos{2.19}} & \makecell{72.90 \pos{0.80}} & \makecell{71.90 \pos{1.89}} \\
        \textbf{OmniJigsaw (JMI)} & \makecell{66.83 \pos{1.95}} & \makecell{62.78 \pos{0.21}} & \makecell{71.08 \pos{0.45}} & \makecell{44.70 \pos{0.40}} & \makecell{50.24 \negdelta{0.60}} & \makecell{58.76 \pos{0.79}} & \makecell{72.90 \pos{0.80}} & \makecell{71.39 \pos{1.38}} \\
        \textbf{OmniJigsaw (SMS)} & \makecell{68.12 \pos{3.24}} & \makecell{65.15 \pos{2.58}} & \makecell{72.03 \pos{1.40}} & \makecell{45.80 \pos{1.50}} & \makecell{52.26 \pos{1.42}} & \makecell{61.75 \pos{3.78}} & \makecell{72.90 \pos{0.80}} & \makecell{\textbf{72.63} \pos{2.62}} \\
        \textbf{OmniJigsaw (CMM)} & \makecell{\textbf{68.90} \pos{4.02}} & \makecell{\textbf{65.29} \pos{2.72}} & \makecell{72.03 \pos{1.40}} & \makecell{\textbf{46.10} \pos{1.80}} & \makecell{\textbf{52.53} \pos{1.69}} & \makecell{\textbf{62.35} \pos{4.38}} & \makecell{\textbf{73.10} \pos{1.00}} & \makecell{72.26 \pos{2.25}} \\
        \midrule
        \multicolumn{9}{c}{\textit{Infer w/o Audio}} \\
        Video-R1        & 52.60 & 55.94 & 70.00 & 40.60 & 42.13 & 52.39 & 62.80 & 67.07 \\
        Omni-R1         & 51.03 & 52.65 & 62.78 & 37.00 & 38.76 & 51.00 & 60.30 & 64.86 \\
        HumanOmniV2     & 47.35 & 49.44 & 63.10 & 38.20 & 38.87 & 48.21 & 61.20 & 66.65 \\
        \rowcolor{gray!15} Qwen3-Omni-30B & 63.32 & 63.62 & 70.70 & 43.80 & 46.60 & 58.76 & 67.90 & 70.98 \\
        \midrule
        VideoJigsaw     & \makecell{66.22 \pos{2.90}} & \makecell{64.39 \pos{0.77}} & \makecell{71.52 \pos{0.82}} & \makecell{44.90 \pos{1.10}} & \makecell{47.90 \pos{1.30}} & \makecell{61.55 \pos{2.79}} & \makecell{68.90 \pos{1.00}} & \makecell{73.14 \pos{2.16}} \\
        \textbf{OmniJigsaw (JMI)} & \makecell{65.83 \pos{2.51}} & \makecell{63.90 \pos{0.28}} & \makecell{71.39 \pos{0.69}} & \makecell{44.80 \pos{1.00}} & \makecell{47.47 \pos{0.87}} & \makecell{59.76 \pos{1.00}} & \makecell{68.80 \pos{0.90}} & \makecell{72.45 \pos{1.47}} \\
        \textbf{OmniJigsaw (SMS)} & \makecell{66.39 \pos{3.07}} & \makecell{64.73 \pos{1.11}} & \makecell{71.52 \pos{0.82}} & \makecell{45.90 \pos{2.10}} & \makecell{47.96 \pos{1.36}} & \makecell{60.76 \pos{2.00}} & \makecell{69.00 \pos{1.10}} & \makecell{72.17 \pos{1.19}} \\
        \textbf{OmniJigsaw (CMM)} & \makecell{\textbf{66.83} \pos{3.51}} & \makecell{\textbf{66.20} \pos{2.58}} & \makecell{\textbf{72.34} \pos{1.64}} & \makecell{\textbf{46.50} \pos{2.70}} & \makecell{\textbf{48.29} \pos{1.69}} & \makecell{\textbf{62.75} \pos{3.99}} & \makecell{\textbf{69.30} \pos{1.40}} & \makecell{\textbf{73.46} \pos{2.48}} \\
        \bottomrule[1.2pt]
    \end{tabular}
\end{table}

\subsection{Implementation Details}
\label{sec:imp_details}

We employ Qwen3-Omni-30B-A3B-Instruct~\cite{xu2025qwen3} as the baseline for GRPO~\cite{shao2024deepseekmath} post-training under the proposed OmniJigsaw framework. The training data (denoted as OmniJigsaw-8K) are curated from YouCook2~\cite{zhou2018towards}, FineVideo~\cite{Farré2024FineVideo}, and LLaVA-Video-178K~\cite{zhang2024llava} using our two-stage filtering pipeline with Qwen2.5-VL-7B-Instruct~\cite{bai2025qwen25vltechnicalreport} as the semantic assessor. Training spans 1000 steps with 6 clips per sample, while the vision tower, audio tower, and router are frozen for efficiency. To investigate modality-specific enhancements, we establish uni-modal VideoJigsaw and AudioJigsaw baselines requiring the model to reorder shuffled clips using exclusively visual or auditory cues. Evaluation is conducted employing greedy decoding with explicit CoT reasoning. Detailed hyperparameters, data partitions, prompts, and uni-modal jigsaw formulations are provided in the Appendix~\ref{sec:appendix}.

\subsection{Main Results}
\label{sec:main_results}

\subsubsection{Video Reasoning}
\label{sec:video_reasoning}

We evaluate video understanding across eight benchmarks spanning foundational temporal sensitivity (AoTBench~\cite{xue2025seeing}, TempCompass~\cite{liu2024tempcompass}, TUNA-Bench~\cite{kong2025tuna}), high-level cognitive reasoning (Video-Holmes~\cite{cheng2025video}, Video-TT~\cite{zhang2025towards}), and holistic multi-task comprehension (Video-MME~\cite{fu2025video}, MLVU~\cite{zhou2025mlvu}, MLVU-Test~\cite{zhou2025mlvu}). Beyond the Qwen3-Omni-30B-A3B-Instruct~\cite{xu2025qwen3} baseline, we further assess Omni-R1~\cite{zhong2025omni}, HumanOmniV2~\cite{yang2025humanomniv2}, and Video-R1~\cite{feng2025video}. Experiments employ two inference modes (\textit{w/ audio} and \textit{w/o audio}), with videos downsampled to 200 frames ($128 \times 28 \times 28$ pixels) and audio symmetrically truncated to 600s. As presented in Table~\ref{tab:video_table}, OmniJigsaw yields substantial gains (\textbf{+4.38} on MLVU-Test) across nearly all benchmarks, with CMM consistently outperforming VideoJigsaw despite auxiliary audio attention allocation. Notable gains on AoTBench (\textbf{+4.02}) and Video-TT (\textbf{+2.70}) indicate that OmniJigsaw profoundly bolsters temporal awareness and active clue association essential for complex reasoning, while exceptional results on MLVU (\textbf{+2.62}) and Video-MME (\textbf{+1.40}) suggest these enhancements facilitate long-range causal inference and global semantic synthesis.

\subsubsection{Audio Reasoning}
\label{sec:audio_reasoning}
\begin{wraptable}{r}{0.53\textwidth}
    \centering
    \vspace{-12pt}
    \caption{\textbf{Performance comparison on audio reasoning.} \textbf{Bold}: best; {\color{green!60!black}green}: gains over baseline.}
    \label{tab:audio_table}
    \fontfamily{ptm}\selectfont
    \renewcommand{\arraystretch}{1.2}
    \setlength{\tabcolsep}{1.8pt}
    \footnotesize
    \begin{tabular}{@{} l *{4}{c} @{}}
        \toprule
        \textbf{Method} &
        \rotatebox{60}{\shortstack[c]{\textbf{\scriptsize MMAU}\\\textbf{\scriptsize -Pro}}} & 
        \rotatebox{60}{\shortstack[c]{\textbf{\scriptsize MMAU}\\\textbf{\scriptsize -test-mini}}} & 
        \rotatebox{60}{\textbf{\scriptsize MMSU}} & 
        \rotatebox{60}{\textbf{\scriptsize MMAR}} \\
        \midrule
        Omni-R1         & 52.36 & \textbf{77.70} & 61.87 & 59.70 \\
        HumanOmniV2     & 53.49 & 75.90 & 60.83 & 61.80 \\
        \rowcolor{gray!15} Qwen3-Omni-30B & 56.61 & 74.40 & 70.16 & 68.50 \\
        \midrule
        AudioJigsaw     & \makecell{57.67 \pos{1.06}} & \makecell{75.40 \pos{1.00}} & \makecell{70.30 \pos{0.14}} & \makecell{70.70 \pos{2.20}} \\
        \textbf{OmniJigsaw (JMI)} & \makecell{58.33 \pos{1.72}} & \makecell{74.50 \pos{0.10}} & \makecell{69.80 \negdelta{0.36}} & \makecell{69.10 \pos{0.60}} \\
        \textbf{OmniJigsaw (SMS)} & \makecell{58.46 \pos{1.85}} & \makecell{75.80 \pos{1.40}} & \makecell{70.48 \pos{0.32}} & \makecell{69.50 \pos{1.00}} \\
        \textbf{OmniJigsaw (CMM)} & \makecell{\textbf{58.59} \pos{1.98}} & \makecell{76.30 \pos{1.90}} & \makecell{\textbf{70.70} \pos{0.54}} & \makecell{\textbf{71.00} \pos{2.50}} \\
        \bottomrule
    \end{tabular}
    \vspace{-20pt}
\end{wraptable}

To evaluate audio understanding improvements facilitated by our OmniJigsaw, we employ four representative benchmarks: MMSU~\cite{wang2025mmsu} for fine-grained perception, MMAU-test-mini~\cite{sakshi2024mmau} and MMAR~\cite{ma2025mmar} for hierarchical reasoning, and MMAU-Pro~\cite{kumar2025mmau} for versatile auditory comprehension. As shown in Table~\ref{tab:audio_table}, OmniJigsaw yields consistent improvements; notably, CMM outperforms AudioJigsaw despite the latter's exclusive audio attention, validating its efficacy in excavating mutually beneficial audio-visual synergy. Significant gains on MMAR (\textbf{+2.50}) and robust performance on MMAU-Pro (\textbf{+1.98}) reflect enhanced structural reasoning and global temporal perception, confirming that OmniJigsaw fosters core acoustic logic and contextual coherence.

\subsubsection{Omni-Modal Collaborative Reasoning}
\label{sec:omni_reasoning}

We evaluate audio-visual collaborative reasoning across three comprehensive omni-modal benchmarks: Daily-Omni~\cite{zhou2025daily} for temporal event synchronization, OmniVideoBench~\cite{li2025omnivideobench} for logical challenges requiring joint omni-modal clue exploitation, and IntentBench~\cite{yang2025humanomniv2} for advanced behavioral and intent inference. As reported in Table~\ref{tab:omni_table}, the universal performance gains of OmniJigsaw across all benchmarks substantiate its profound omni-modal empowerment effect. Gains yielded by JMI and SMS on \begin{wraptable}{r}{0.47\textwidth}
    \centering
    \vspace{-32pt}
    \caption{\textbf{Performance comparison on omni-modal collaborative reasoning.} \textbf{Bold}: best; {\color{green!60!black}green}: gains over baseline.}
    \label{tab:omni_table}
    \fontfamily{ptm}\selectfont
    \renewcommand{\arraystretch}{1.2}
    \setlength{\tabcolsep}{1.8pt}
    \footnotesize
    \begin{tabular}{@{} l *{3}{c} @{}}
        \toprule[1.2pt]
        \textbf{Method} &
        \rotatebox{60}{\shortstack[c]{\textbf{\scriptsize Daily}\\\textbf{\scriptsize -Omni}}} & 
        \rotatebox{60}{\shortstack[c]{\textbf{\scriptsize Intent}\\\textbf{\scriptsize Bench}}} & 
        \rotatebox{60}{\shortstack[c]{\textbf{\scriptsize Omni}\\\textbf{\scriptsize Video}\\\textbf{\scriptsize Bench}}} \\
        \midrule
        
        Omni-R1         & 54.14 & 64.18 & 32.40 \\
        HumanOmniV2     & 58.48 & 68.21 & 33.50 \\
        \rowcolor{gray!15} Qwen3-Omni-30B & 69.92 & 67.40 & 38.80 \\
        \midrule
        
        \textbf{OmniJigsaw (JMI)} & \makecell{70.26 \pos{0.34}} & \makecell{67.95 \pos{0.55}} & \makecell{40.10 \pos{1.30}} \\
        
        \textbf{OmniJigsaw (SMS)} & \makecell{70.26 \pos{0.34}} & \makecell{68.21 \pos{0.81}} & \makecell{40.20 \pos{1.40}} \\

        \textbf{OmniJigsaw (CMM)} & \makecell{\textbf{71.09} \pos{1.17}} & \makecell{\textbf{68.89} \pos{1.49}} & \makecell{\textbf{40.50} \pos{1.70}} \\
        
        \bottomrule[1.2pt]
    \end{tabular}
    \vspace{-20pt}
\end{wraptable}IntentBench reflect enhanced behavioral perception and progress in strategically leveraging salient intent-driven modal cues, while the robust performance of our CMM on OmniVideoBench (\textbf{+1.70}) confirms that enforcing cross-modal dependencies effectively awakens a synergistic perception of complementary cues. By employing these specialized modality orchestration strategies that target modality arbitration and cross-modal dependency modeling, OmniJigsaw facilitates a critical transition from discrete signal perception to unified reasoning logic, thereby significantly strengthening the model’s modality representation cohesion during collaborative reasoning. More qualitative examples are shown in the Appendix~\ref{sec:appendix}.

\subsection{Ablations and Analysis}
\label{sec:ablations_analysis}

\begin{table}[!t]
    \centering
    \caption{\textbf{Ablation study on data quality and reward discount factor} across video, audio, and omni-modal benchmarks under CMM.}
    \label{tab:combined_ablations}
    \fontfamily{ptm}\selectfont
    \renewcommand{\arraystretch}{1.2} 
    \setlength{\tabcolsep}{1.8pt}
    
    \resizebox{\textwidth}{!}{
    \begin{tabular}{@{} l | cccccccc | cccc | ccc @{}}
        \toprule[1.2pt]
        & \multicolumn{8}{c|}{\textbf{Video}} & \multicolumn{4}{c|}{\textbf{Audio}} & \multicolumn{3}{c}{\textbf{Omni-Modal}} \\
        \cmidrule(lr){2-9} \cmidrule(lr){10-13} \cmidrule(l){14-16}
        \textbf{Method} &
        \rotatebox{60}{\textbf{\scriptsize AoT}} & 
        \rotatebox{60}{\textbf{\scriptsize TUNA}} & 
        \rotatebox{60}{\textbf{\scriptsize TempC}} & 
        \rotatebox{60}{\textbf{\scriptsize V-TT}} & 
        \rotatebox{60}{\textbf{\scriptsize V-Holmes}} & 
        \rotatebox{60}{\textbf{\scriptsize MLVU-T}} & 
        \rotatebox{60}{\textbf{\scriptsize V-MME}} & 
        \rotatebox{60}{\textbf{\scriptsize MLVU}} & 
        \rotatebox{60}{\textbf{\scriptsize MMAU-P}} & 
        \rotatebox{60}{\textbf{\scriptsize MMAU-TM}} & 
        \rotatebox{60}{\textbf{\scriptsize MMSU}} & 
        \rotatebox{60}{\textbf{\scriptsize MMAR}} & 
        \rotatebox{60}{\textbf{\scriptsize D-Omni}} & 
        \rotatebox{60}{\textbf{\scriptsize Intent}} & 
        \rotatebox{60}{\textbf{\scriptsize OVB}} \\
        \midrule
        
        \rowcolor{gray!15} \textbf{OmniJigsaw} & \textbf{66.83} & \textbf{66.20} & \textbf{72.34} & \textbf{46.50} & \textbf{48.29} & \textbf{62.75} & \textbf{69.30} & \textbf{73.46} & \textbf{58.59} & \textbf{76.30} & \textbf{70.70} & \textbf{71.00} & \textbf{71.09} & \textbf{68.89} & \textbf{40.50} \\
        \midrule
        
        w/o Filtering & \makecell{64.94 \dataloss{1.89}} & \makecell{65.29 \dataloss{0.91}} & \makecell{71.14 \dataloss{1.20}} & \makecell{45.40 \dataloss{1.10}} & \makecell{47.41 \dataloss{0.88}} & \makecell{58.76 \dataloss{3.99}} & \makecell{68.60 \dataloss{0.70}} & \makecell{72.59 \dataloss{0.87}} & \makecell{57.67 \dataloss{0.92}} & \makecell{76.00 \dataloss{0.30}} & \makecell{70.40 \dataloss{0.30}} & \makecell{68.90 \dataloss{2.10}} & \makecell{70.01 \dataloss{1.08}} & \makecell{66.77 \dataloss{2.12}} & \makecell{40.10 \dataloss{0.40}} \\
        
        w/o DF & \makecell{66.00 \mecloss{0.83}} & \makecell{64.11 \mecloss{2.09}} & \makecell{71.27 \mecloss{1.07}} & \makecell{44.70 \mecloss{1.80}} & \makecell{47.96 \mecloss{0.33}} & \makecell{61.95 \mecloss{0.80}} & \makecell{68.60 \mecloss{0.70}} & \makecell{72.36 \mecloss{1.10}} & \makecell{57.81 \mecloss{0.78}} & \makecell{75.90 \mecloss{0.40}} & \makecell{70.48 \mecloss{0.22}} & \makecell{69.30 \mecloss{1.70}} & \makecell{69.76 \mecloss{1.33}} & \makecell{67.66 \mecloss{1.23}} & \makecell{39.50 \mecloss{1.00}} \\
        
        \bottomrule[1.2pt]
    \end{tabular}
    }
\end{table}

\begin{figure}[t]
    \centering
    \includegraphics[width=\linewidth]{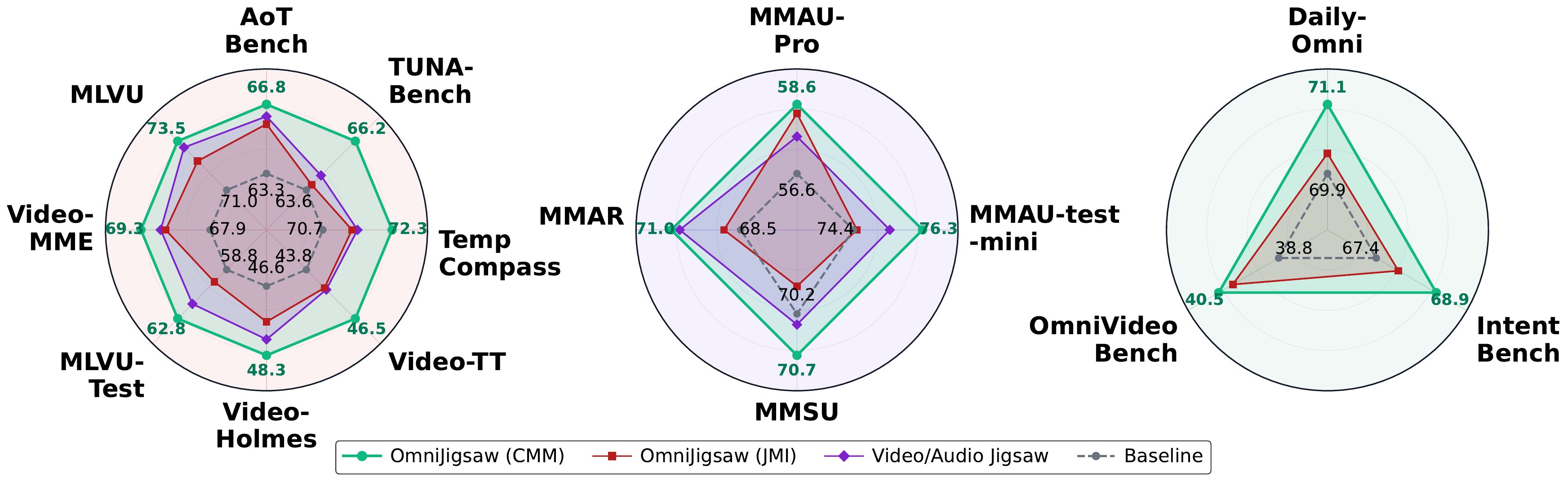}
    \caption{\textbf{Performance comparison of JMI, CMM, and uni-modal Jigsaw across video, audio, and omni-modal benchmarks.} CMM's consistent superiority and JMI's performance degradation relative to uni-modal Jigsaw baselines compellingly support the ``bi-modal shortcut phenomenon''.}
    \label{fig:radar_compare}
\end{figure}

\subsubsection{Sensitivity to Data Quality}
\label{sec:data_ablations}

To quantify the impact of data quality on OmniJigsaw post-training efficacy, we construct a control group by randomly sampling a subset of equivalent size from LLaVA-Video-178K~\cite{zhang2024llava} for RL post-training under the best-performing CMM strategy without filtration. As shown in Table~\ref{tab:combined_ablations}, the model trained utilizing random data performs consistently worse (\textbf{-3.99} on MLVU-Test~\cite{zhou2025mlvu}; \textbf{-2.10} on MMAR~\cite{ma2025mmar}; \textbf{-2.12} on IntentBench~\cite{yang2025humanomniv2}) than the OmniJigsaw-8K group. This performance disparity corroborates the importance of data quality to the success of OmniJigsaw. Fundamentally, the solvability of reordering tasks is predicated on identifiable state evolution between clips. If the original samples lack dynamism (\eg, quasi-static videos), the generated clips will exhibit high visual redundancy, resulting in a loss of definitive logical chronological association due to the absence of discriminative differences. Such theoretically ill-posed samples fail to provide effective supervision signals, thereby hindering the model from establishing robust cross-modal temporal representations. By systematically eliminating these pathological samples, our filtering pipeline provides scalable support for the efficient adaptation of OmniJigsaw to massive unannotated omni-modal data, thereby catalyzing the emergence of more robust and capable omni-modal models.

\begin{figure}[t]
    \centering
    \includegraphics[width=\linewidth]{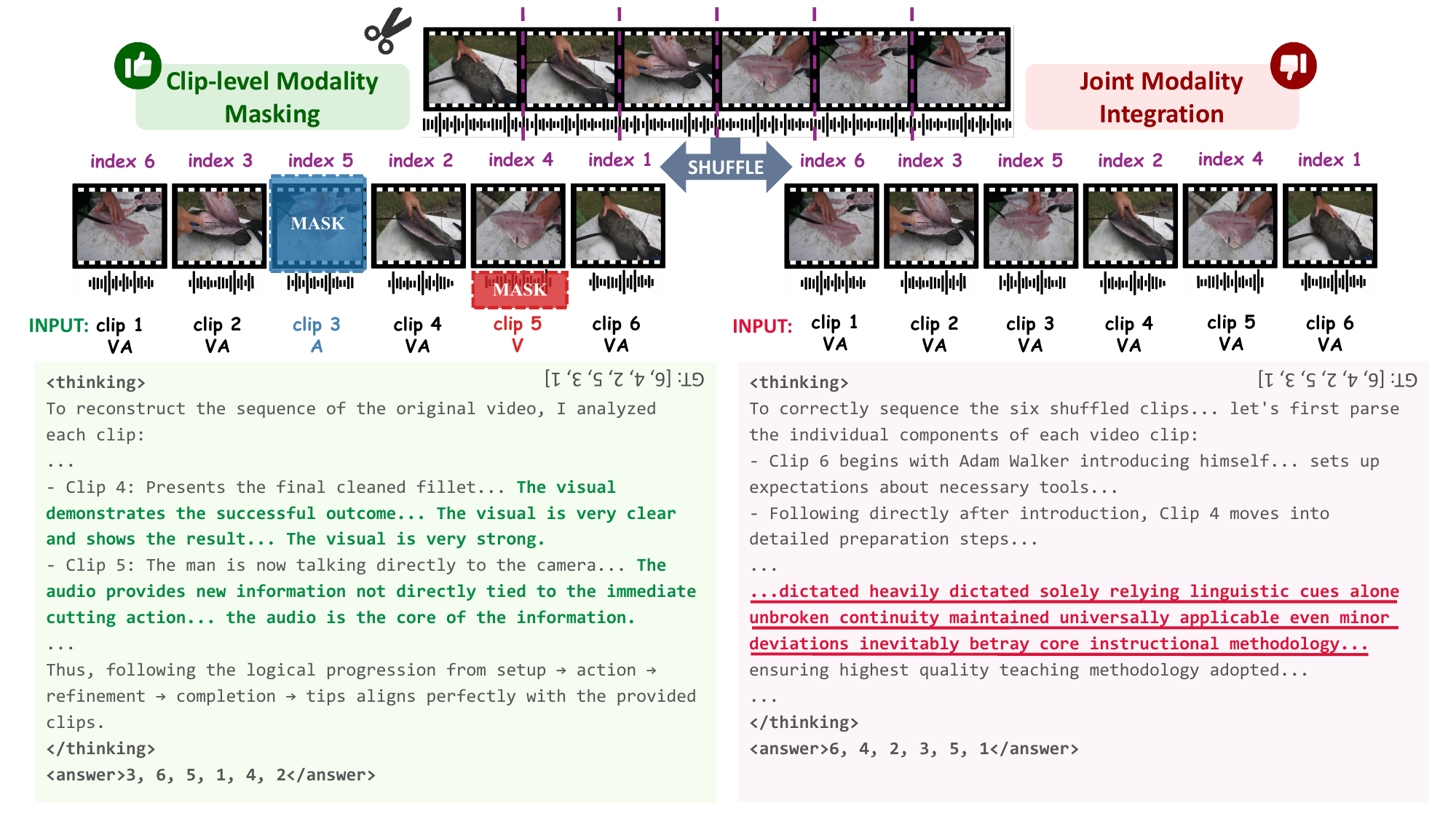}
    \caption{\textbf{Comparison of CoT reasoning between CMM and JMI at training step 800.} CMM (left) compels the model to jointly analyze visual and auditory cues by masking less salient modalities (dashed boxes) to create an information bottleneck, while JMI (right) exhibits a bi-modal shortcut by \textbf{``solely relying [on] linguistic cues''} (underlined text) and bypassing the necessary visual analysis.}
    \label{fig:rollout_compare}
\end{figure}

\subsubsection{Discount Factor as a Catalyst}
\label{sec:df_ablations}

\begin{wrapfigure}{r}{0.4\textwidth}
    \centering
    \vspace{-22pt}
    \includegraphics[width=\linewidth]{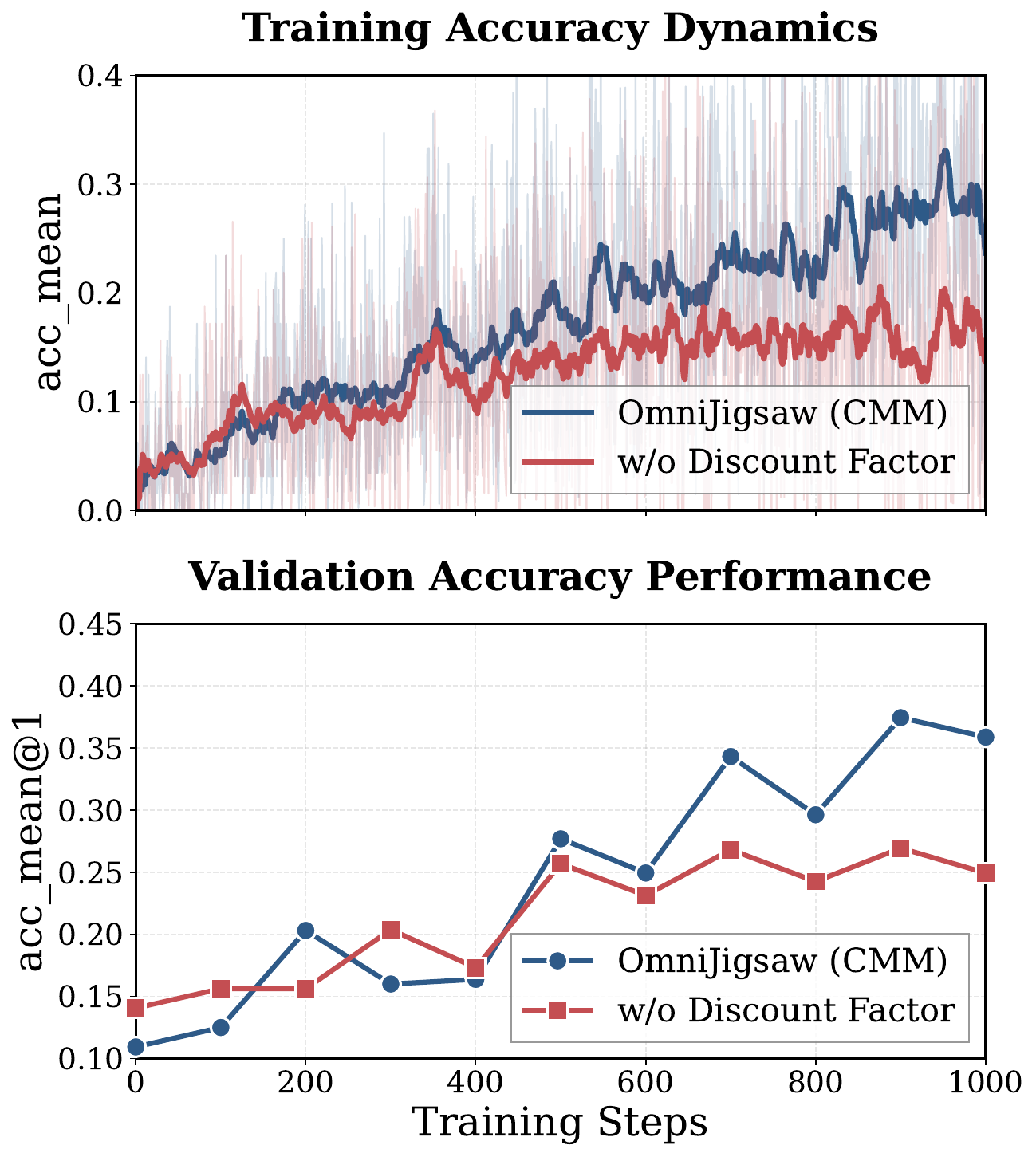}
    \vspace{-12pt}
    \caption{\textbf{Optimization dynamics w/ and w/o Discount Factor.}}
    \label{fig:df_ablation}
    \vspace{-22pt}
\end{wrapfigure}

To analyze the impact of the discount factor in reward design, we likewise instantiate an ablation variant based on the superior CMM strategy, where the discount factor $\lambda$ is fixed at $1$. The training and validation \texttt{acc\_mean} curves in Fig.~\ref{fig:df_ablation} illustrate that the inclusion of the discount factor facilitates a persistent upward trajectory by fostering superior exploration, while its absence causes the model to converge prematurely at a sub-optimal level. These observations are further corroborated by the performance degradation (\textbf{-2.09} on TUNA-Bench~\cite{kong2025tuna}; \textbf{-1.70} on MMAR~\cite{ma2025mmar}; \textbf{-1.33} on Daily-Omni~\cite{zhou2025daily}) recorded in Table~\ref{tab:combined_ablations}. Mechanistically, by suppressing reward weights for imperfect sequences, this design artificially amplifies the value disparity between ``sub-optimal'' and ``optimal'' solutions within the GRPO group, which drives the model to explore optimization spaces more aggressively in later training stages, thereby effectively circumventing the risk of entrapment in non-optimal plateaus.

\subsubsection{Are More Modalities Always Better?}
\label{sec:bimodal_analysis}

As indicated in Tables~\ref{tab:video_table} and~\ref{tab:audio_table}, independent VideoJigsaw and AudioJigsaw effectively enhance the understanding capabilities of their respective modalities. However, does the integration of a second modality inherently lead to further performance gains? Our research suggests that the answer hinges on the inter-modal interaction and utilization mechanisms within the proxy task. Regarding JMI, as depicted in Fig.~\ref{fig:radar_compare}, we observe a counter-intuitive phenomenon where it underperforms uni-modal Jigsaw baselines for both video and audio reasoning. We denote this phenomenon\begin{wrapfigure}{r}{0.5\textwidth}
    \centering
    \vspace{-20pt}
    \includegraphics[width=\linewidth]{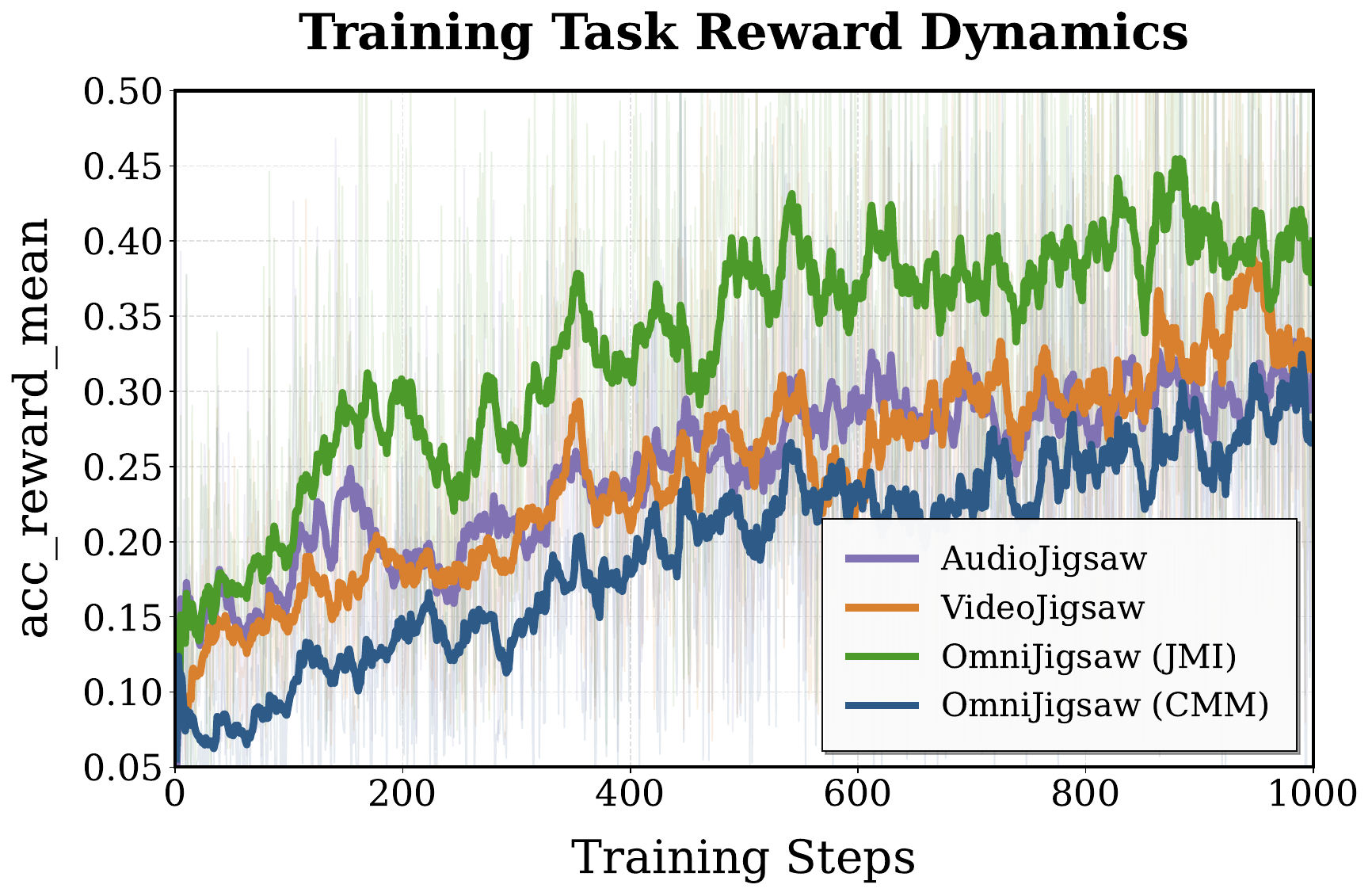}
    \vspace{-10pt}
    \caption{\textbf{Task reward (w/o $R_{rep}, R_{fmt}$) dynamics of JMI, CMM, VideoJigsaw and AudioJigsaw.}}
    \label{fig:reward_compare}
    \vspace{-25pt}
\end{wrapfigure} within OmniJigsaw as the ``bi-modal shortcut phenomenon''. Under JMI, the complete audio-visual stream provides redundant solution paths, encouraging the model to preferentially rely on the information-rich dominant modality to derive answers based on sample characteristics, thus bypassing deep analysis of the other modality. As shown in Figs.~\ref{fig:rollout_compare} and~\ref{fig:reward_compare}, although redundant bi-modal cues reduce task difficulty (evidenced by the shortcut reasoning patterns within the CoT and the significantly higher \texttt{acc\_reward\_mean}), this facile victory attenuates the necessity to mine cues from the weaker modality, leading to insufficient representation learning. In contrast, our two advanced strategies (CMM and SMS) effectively mitigate this effect. CMM introduces an information bottleneck via dynamic clip-level shielding, blocking the possibility of completing tasks via single-modality reliance. This compels the model to perform deep switching and integration between video and audio cues, transforming inter-modal ``short-circuiting'' into ``mutual synergy'', thereby achieving performance that surpasses uni-modal baselines. Similarly, SMS employs a sample-level preference mechanism that locks onto the optimal temporal modality while effectively filtering non-dominant modal noise, also yielding training performance superior to VideoJigsaw and AudioJigsaw.

\subsubsection{Sample-level Arbitration or Clip-level Orchestration?}
\label{sec:granularity_analysis}

\begin{figure}[t]
    \centering
    \includegraphics[width=\linewidth]{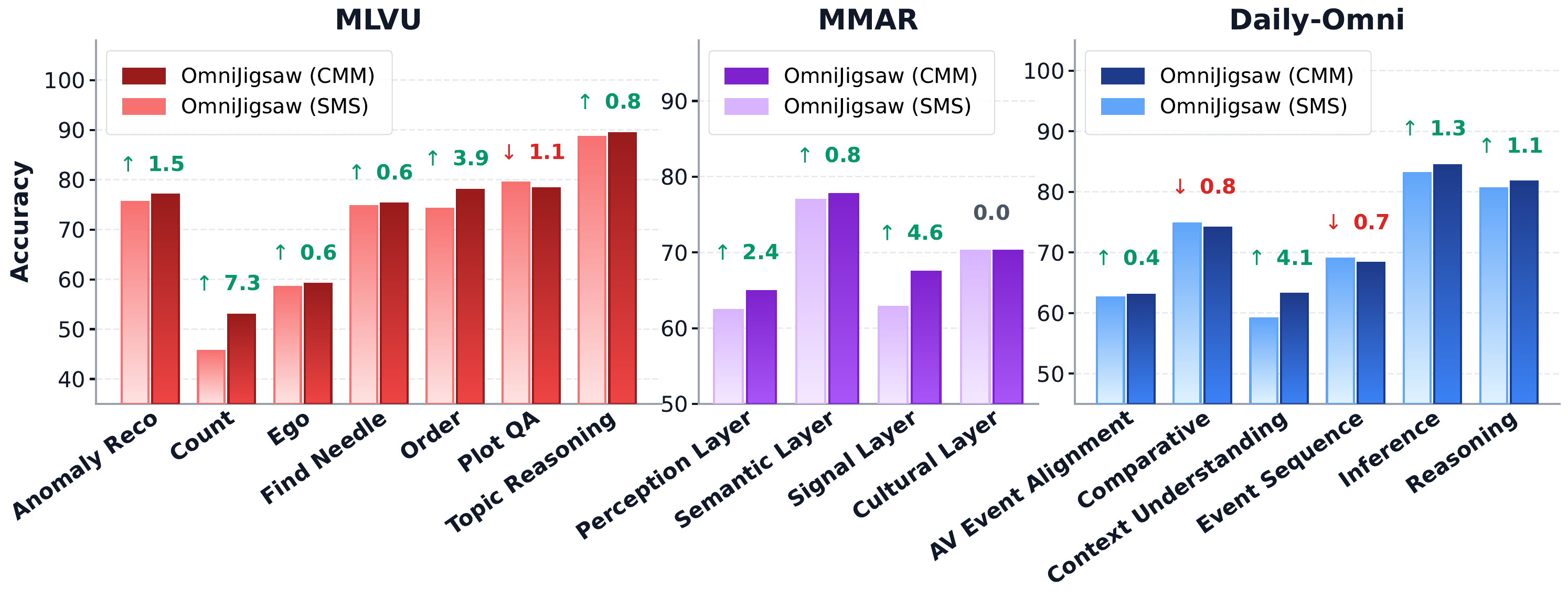}
    \caption{\textbf{Sub-capability performance comparison between CMM and SMS across fine-grained dimensions.} CMM's predominant superiority over SMS highlights the efficacy of clip-level orchestration in capturing temporally non-uniform audio-visual cues, whereas sample-level arbitration often misses local high-value modal information.}
    \label{fig:bar_compare}
\end{figure}

Figure~\ref{fig:bar_compare} provides a detailed comparison of CMM and SMS across fine-grained sub-capability dimensions covering representative video, audio, and omni-modal benchmarks (MLVU~\cite{zhou2025mlvu}, MMAR~\cite{ma2025mmar}, and Daily-Omni~\cite{zhou2025daily}). CMM consistently outperforms SMS across almost all dimensions. This performance disparity highlights the notable effect of granularity in data-adaptive strategies. Audio-visual cues in real-world scenarios typically exhibit characteristics where the dominant modality alternates non-uniformly along the temporal axis. Although the sample-level global arbitration adopted by SMS prioritizes the modality with a higher overall signal-to-noise ratio, it potentially leads to the omission of local high-value modal cues. In contrast, CMM executes a fine-grained local modality selection that conforms to the dynamic flow of audio-visual information, which not only maximizes local information entropy but also compels the model to perform cross-modal semantic stitching among fragmented heterogeneous cues, essentially deepening the model's analytical insight for complex multimodal scenarios.

\section{Conclusion}
\label{sec:conclusion}
In this work, we present OmniJigsaw, a scalable self-supervised RL post-training framework designed to enhance omni-modal reasoning by orchestrating audio-visual signals through three distinct strategies within a temporal reordering proxy task. Extensive evaluations across 15 benchmarks demonstrate that OmniJigsaw yields substantial improvements in video, audio, and omni-modal collaborative reasoning. To ensure scalability, we establish a two-stage coarse-to-fine data filtering pipeline that supports the efficient adaptation of OmniJigsaw to massive unannotated omni-modal data. Furthermore, comprehensive analysis reveals that our fine-grained Clip-level Modality Masking strategy effectively mitigates the inherent ``bi-modal shortcut phenomenon'' caused by redundant modality participation. Ultimately, OmniJigsaw highlights the profound potential of a lightweight, annotation-free post-training paradigm in cultivating omni-modal models with advanced complex reasoning capabilities.

\bibliographystyle{splncs04}
\bibliography{main}

@article{ouyang2022training,
  title={Training language models to follow instructions with human feedback},
  author={Ouyang, Long and Wu, Jeffrey and Jiang, Xu and Almeida, Diogo and Wainwright, Carroll and Mishkin, Pamela and Zhang, Chong and Agarwal, Sandhini and Slama, Katarina and Ray, Alex and others},
  journal={Advances in neural information processing systems},
  volume={35},
  pages={27730--27744},
  year={2022}
}

@article{rafailov2023direct,
  title={Direct preference optimization: Your language model is secretly a reward model},
  author={Rafailov, Rafael and Sharma, Archit and Mitchell, Eric and Manning, Christopher D and Ermon, Stefano and Finn, Chelsea},
  journal={Advances in neural information processing systems},
  volume={36},
  pages={53728--53741},
  year={2023}
}

@article{lambert2024tulu,
  title={Tulu 3: Pushing frontiers in open language model post-training},
  author={Lambert, Nathan and Morrison, Jacob and Pyatkin, Valentina and Huang, Shengyi and Ivison, Hamish and Brahman, Faeze and Miranda, Lester James V and Liu, Alisa and Dziri, Nouha and Lyu, Shane and others},
  journal={arXiv preprint arXiv:2411.15124},
  year={2024}
}

@article{guo2025deepseek,
  title={Deepseek-r1: Incentivizing reasoning capability in llms via reinforcement learning},
  author={Guo, Daya and Yang, Dejian and Zhang, Haowei and Song, Junxiao and Wang, Peiyi and Zhu, Qihao and Xu, Runxin and Zhang, Ruoyu and Ma, Shirong and Bi, Xiao and others},
  journal={arXiv preprint arXiv:2501.12948},
  year={2025}
}

@article{roziere2023code,
  title={Code llama: Open foundation models for code},
  author={Roziere, Baptiste and Gehring, Jonas and Gloeckle, Fabian and Sootla, Sten and Gat, Itai and Tan, Xiaoqing Ellen and Adi, Yossi and Liu, Jingyu and Sauvestre, Romain and Remez, Tal and others},
  journal={arXiv preprint arXiv:2308.12950},
  year={2023}
}

@article{li2025videochat,
  title={Videochat-r1: Enhancing spatio-temporal perception via reinforcement fine-tuning},
  author={Li, Xinhao and Yan, Ziang and Meng, Desen and Dong, Lu and Zeng, Xiangyu and He, Yinan and Wang, Yali and Qiao, Yu and Wang, Yi and Wang, Limin},
  journal={arXiv preprint arXiv:2504.06958},
  year={2025}
}

@misc{xu2025qwen25omnitechnicalreport,
      title={Qwen2.5-Omni Technical Report}, 
      author={Jin Xu and Zhifang Guo and Jinzheng He and Hangrui Hu and Ting He and Shuai Bai and Keqin Chen and Jialin Wang and Yang Fan and Kai Dang and Bin Zhang and Xiong Wang and Yunfei Chu and Junyang Lin},
      year={2025},
      eprint={2503.20215},
      archivePrefix={arXiv},
      primaryClass={cs.CL},
      url={https://arxiv.org/abs/2503.20215}, 
}

@article{xu2025qwen3,
  title={Qwen3-omni technical report},
  author={Xu, Jin and Guo, Zhifang and Hu, Hangrui and Chu, Yunfei and Wang, Xiong and He, Jinzheng and Wang, Yuxuan and Shi, Xian and He, Ting and Zhu, Xinfa and others},
  journal={arXiv preprint arXiv:2509.17765},
  year={2025}
}

@article{grattafiori2024llama,
  title={The llama 3 herd of models},
  author={Grattafiori, Aaron and Dubey, Abhimanyu and Jauhri, Abhinav and Pandey, Abhinav and Kadian, Abhishek and Al-Dahle, Ahmad and Letman, Aiesha and Mathur, Akhil and Schelten, Alan and Vaughan, Alex and others},
  journal={arXiv preprint arXiv:2407.21783},
  year={2024}
}

@article{yang2024qwen2,
  title={Qwen2. 5-math technical report: Toward mathematical expert model via self-improvement},
  author={Yang, An and Zhang, Beichen and Hui, Binyuan and Gao, Bofei and Yu, Bowen and Li, Chengpeng and Liu, Dayiheng and Tu, Jianhong and Zhou, Jingren and Lin, Junyang and others},
  journal={arXiv preprint arXiv:2409.12122},
  year={2024}
}

@article{tunstall2023zephyr,
  title={Zephyr: Direct distillation of lm alignment},
  author={Tunstall, Lewis and Beeching, Edward and Lambert, Nathan and Rajani, Nazneen and Rasul, Kashif and Belkada, Younes and Huang, Shengyi and Von Werra, Leandro and Fourrier, Cl{\'e}mentine and Habib, Nathan and others},
  journal={arXiv preprint arXiv:2310.16944},
  year={2023}
}

@article{zhu2024deepseek,
  title={Deepseek-coder-v2: Breaking the barrier of closed-source models in code intelligence},
  author={Zhu, Qihao and Guo, Daya and Shao, Zhihong and Yang, Dejian and Wang, Peiyi and Xu, Runxin and Wu, Y and Li, Yukun and Gao, Huazuo and Ma, Shirong and others},
  journal={arXiv preprint arXiv:2406.11931},
  year={2024}
}

@inproceedings{yu2024rlhf,
  title={Rlhf-v: Towards trustworthy mllms via behavior alignment from fine-grained correctional human feedback},
  author={Yu, Tianyu and Yao, Yuan and Zhang, Haoye and He, Taiwen and Han, Yifeng and Cui, Ganqu and Hu, Jinyi and Liu, Zhiyuan and Zheng, Hai-Tao and Sun, Maosong and others},
  booktitle={Proceedings of the IEEE/CVF Conference on Computer Vision and Pattern Recognition},
  pages={13807--13816},
  year={2024}
}

@inproceedings{wallace2024diffusion,
  title={Diffusion model alignment using direct preference optimization},
  author={Wallace, Bram and Dang, Meihua and Rafailov, Rafael and Zhou, Linqi and Lou, Aaron and Purushwalkam, Senthil and Ermon, Stefano and Xiong, Caiming and Joty, Shafiq and Naik, Nikhil},
  booktitle={Proceedings of the IEEE/CVF Conference on Computer Vision and Pattern Recognition},
  pages={8228--8238},
  year={2024}
}

@inproceedings{liu2025visual,
  title={Visual-rft: Visual reinforcement fine-tuning},
  author={Liu, Ziyu and Sun, Zeyi and Zang, Yuhang and Dong, Xiaoyi and Cao, Yuhang and Duan, Haodong and Lin, Dahua and Wang, Jiaqi},
  booktitle={Proceedings of the IEEE/CVF International Conference on Computer Vision},
  pages={2034--2044},
  year={2025}
}

@article{chen2026omnivideo,
  title={OmniVideo-R1: Reinforcing Audio-visual Reasoning with Query Intention and Modality Attention},
  author={Chen, Zhangquan and Tao, Jiale and Li, Ruihuang and Hu, Yihao and Chen, Ruitao and Yang, Zhantao and Yu, Xinlei and Jing, Haodong and Zhang, Manyuan and Shao, Shuai and others},
  journal={arXiv preprint arXiv:2602.05847},
  year={2026}
}

@article{li2025comprehensive,
  title={A comprehensive survey on world models for embodied ai},
  author={Li, Xinqing and He, Xin and Zhang, Le and Wu, Min and Li, Xiaoli and Liu, Yun},
  journal={arXiv preprint arXiv:2510.16732},
  year={2025}
}

@article{yue2025simulating,
  title={Simulating the Visual World with Artificial Intelligence: A Roadmap},
  author={Yue, Jingtong and Huang, Ziqi and Chen, Zhaoxi and Wang, Xintao and Wan, Pengfei and Liu, Ziwei},
  journal={arXiv preprint arXiv:2511.08585},
  year={2025}
}

@article{zhou2025reinforced,
  title={Reinforced mllm: A survey on rl-based reasoning in multimodal large language models},
  author={Zhou, Guanghao and Qiu, Panjia and Chen, Cen and Wang, Jie and Yang, Zheming and Xu, Jian and Qiu, Minghui},
  journal={arXiv preprint arXiv:2504.21277},
  year={2025}
}

@article{li2025perception,
  title={Perception, reason, think, and plan: A survey on large multimodal reasoning models},
  author={Li, Yunxin and Liu, Zhenyu and Li, Zitao and Zhang, Xuanyu and Xu, Zhenran and Chen, Xinyu and Shi, Haoyuan and Jiang, Shenyuan and Wang, Xintong and Wang, Jifang and others},
  journal={arXiv preprint arXiv:2505.04921},
  year={2025}
}

@article{wu2025visual,
  title={Visual jigsaw post-training improves mllms},
  author={Wu, Penghao and Zhang, Yushan and Diao, Haiwen and Li, Bo and Lu, Lewei and Liu, Ziwei},
  journal={arXiv preprint arXiv:2509.25190},
  year={2025}
}

@article{wang2025test,
  title={Test-Time Temporal Sampling for Efficient MLLM Video Understanding},
  author={Wang, Kaibin and Lin, Mingbao},
  journal={arXiv preprint arXiv:2511.17945},
  year={2025}
}

@article{tu2025favor,
  title={Favor-bench: A comprehensive benchmark for fine-grained video motion understanding},
  author={Tu, Chongjun and Zhang, Lin and Chen, Pengtao and Ye, Peng and Zeng, Xianfang and Cheng, Wei and Yu, Gang and Chen, Tao},
  journal={arXiv preprint arXiv:2503.14935},
  year={2025}
}

@article{yang2026mllm,
  title={MLLM-VADStory: Domain Knowledge-Driven Multimodal LLMs for Video Ad Storyline Insights},
  author={Yang, Jasmine and Zhang, Poppy and Hill, Shawndra},
  journal={arXiv preprint arXiv:2601.07850},
  year={2026}
}

@article{ccoban2024mllms,
  title={What do MLLMs hear? Examining reasoning with text and sound components in Multimodal Large Language Models},
  author={{\c{C}}oban, Enis Berk and Mandel, Michael I and Devaney, Johanna},
  journal={arXiv preprint arXiv:2406.04615},
  year={2024}
}

@article{yang2025omni,
  title={Omni-emotion: Extending video mllm with detailed face and audio modeling for multimodal emotion analysis},
  author={Yang, Qize and Bai, Detao and Peng, Yi-Xing and Wei, Xihan},
  journal={arXiv preprint arXiv:2501.09502},
  year={2025}
}

@article{guan2025mllm,
  title={MLLM-based Speech Recognition: When and How is Multimodality Beneficial?},
  author={Guan, Yiwen and Trinh, Viet Anh and Voleti, Vivek and Whitehill, Jacob},
  journal={arXiv preprint arXiv:2507.19037},
  year={2025}
}

@misc{ren2026videoworld2learningtransferable,
      title={VideoWorld 2: Learning Transferable Knowledge from Real-world Videos}, 
      author={Zhongwei Ren and Yunchao Wei and Xiao Yu and Guixun Luo and Yao Zhao and Bingyi Kang and Jiashi Feng and Xiaojie Jin},
      year={2026},
      eprint={2602.10102},
      archivePrefix={arXiv},
      primaryClass={cs.CV},
      url={https://arxiv.org/abs/2602.10102}, 
}

@misc{yang2026dualipodualiterativepreferenceoptimization,
      title={Dual-IPO: Dual-Iterative Preference Optimization for Text-to-Video Generation}, 
      author={Xiaomeng Yang and Mengping Yang and Jia Gong and Luozheng Qin and Zhiyu Tan and Hao Li},
      year={2026},
      eprint={2502.02088},
      archivePrefix={arXiv},
      primaryClass={cs.CV},
      url={https://arxiv.org/abs/2502.02088}, 
}

@article{zhao2025unified,
  title={Unified multimodal understanding and generation models: Advances, challenges, and opportunities},
  author={Zhao, Shanshan and Zhang, Xinjie and Guo, Jintao and Hu, Jiakui and Duan, Lunhao and Fu, Minghao and Chng, Yong Xien and Wang, Guo-Hua and Chen, Qing-Guo and Xu, Zhao and others},
  journal={arXiv preprint arXiv:2505.02567},
  year={2025}
}

@inproceedings{zhang2025video,
  title={Video-cot: A comprehensive dataset for spatiotemporal understanding of videos based on chain-of-thought},
  author={Zhang, Shuyi and Hao, Xiaoshuai and Tang, Yingbo and Zhang, Lingfeng and Wang, Pengwei and Wang, Zhongyuan and Ma, Hongxuan and Zhang, Shanghang},
  booktitle={Proceedings of the 33rd ACM International Conference on Multimedia},
  pages={12745--12752},
  year={2025}
}

@article{wang2025cotasks,
  title={CoTasks: Chain-of-Thought based Video Instruction Tuning Tasks},
  author={Wang, Yanan and Vizcarra, Julio and Li, Zhi and Niu, Hao and Kurokawa, Mori},
  journal={arXiv preprint arXiv:2507.13609},
  year={2025}
}

@article{jiang2025videop2r,
  title={VIDEOP2R: Video Understanding from Perception to Reasoning},
  author={Jiang, Yifan and Wang, Yueying and Zhao, Rui and Parag, Toufiq and Chen, Zhimin and Liao, Zhenyu and Unnikrishnan, Jayakrishnan},
  journal={arXiv preprint arXiv:2511.11113},
  year={2025}
}

@inproceedings{noroozi2016unsupervised,
  title={Unsupervised learning of visual representations by solving jigsaw puzzles},
  author={Noroozi, Mehdi and Favaro, Paolo},
  booktitle={European conference on computer vision},
  pages={69--84},
  year={2016},
  organization={Springer}
}

@inproceedings{misra2016shuffle,
  title={Shuffle and learn: unsupervised learning using temporal order verification},
  author={Misra, Ishan and Zitnick, C Lawrence and Hebert, Martial},
  booktitle={European conference on computer vision},
  pages={527--544},
  year={2016},
  organization={Springer}
}

@inproceedings{taleb2021multimodal,
  title={Multimodal self-supervised learning for medical image analysis},
  author={Taleb, Aiham and Lippert, Christoph and Klein, Tassilo and Nabi, Moin},
  booktitle={International conference on information processing in medical imaging},
  pages={661--673},
  year={2021},
  organization={Springer}
}

@article{yang2019xlnet,
  title={Xlnet: Generalized autoregressive pretraining for language understanding},
  author={Yang, Zhilin and Dai, Zihang and Yang, Yiming and Carbonell, Jaime and Salakhutdinov, Russ R and Le, Quoc V},
  journal={Advances in neural information processing systems},
  volume={32},
  year={2019}
}

@article{sauder2019self,
  title={Self-supervised deep learning on point clouds by reconstructing space},
  author={Sauder, Jonathan and Sievers, Bjarne},
  journal={Advances in neural information processing systems},
  volume={32},
  year={2019}
}

@article{xue2025seeing,
  title={Seeing the arrow of time in large multimodal models},
  author={Xue, Zihui and Luo, Mi and Grauman, Kristen},
  journal={arXiv preprint arXiv:2506.03340},
  year={2025}
}

@inproceedings{kong2025tuna,
  title={Tuna: Comprehensive fine-grained temporal understanding evaluation on dense dynamic videos},
  author={Kong, Fanheng and Zhang, Jingyuan and Zhang, Hongzhi and Feng, Shi and Wang, Daling and Yu, Linhao and Ji, Xingguang and Tian, Yu and Zhang, Fuzheng and others},
  booktitle={Proceedings of the 63rd Annual Meeting of the Association for Computational Linguistics (Volume 1: Long Papers)},
  pages={1810--1839},
  year={2025}
}

@inproceedings{liu2024tempcompass,
  title={Tempcompass: Do video llms really understand videos?},
  author={Liu, Yuanxin and Li, Shicheng and Liu, Yi and Wang, Yuxiang and Ren, Shuhuai and Li, Lei and Chen, Sishuo and Sun, Xu and Hou, Lu},
  booktitle={Findings of the Association for Computational Linguistics: ACL 2024},
  pages={8731--8772},
  year={2024}
}

@inproceedings{zhang2025towards,
  title={Towards video thinking test: A holistic benchmark for advanced video reasoning and understanding},
  author={Zhang, Yuanhan and Chew, Yunice and Dong, Yuhao and Leo, Aria and Hu, Bo and Liu, Ziwei},
  booktitle={Proceedings of the IEEE/CVF International Conference on Computer Vision},
  pages={20626--20636},
  year={2025}
}

@article{cheng2025video,
  title={Video-Holmes: Can MLLM Think Like Holmes for Complex Video Reasoning?},
  author={Cheng, Junhao and Ge, Yuying and Wang, Teng and Ge, Yixiao and Liao, Jing and Shan, Ying},
  journal={arXiv preprint arXiv:2505.21374},
  year={2025}
}

@inproceedings{zhou2025mlvu,
  title={Mlvu: Benchmarking multi-task long video understanding},
  author={Zhou, Junjie and Shu, Yan and Zhao, Bo and Wu, Boya and Liang, Zhengyang and Xiao, Shitao and Qin, Minghao and Yang, Xi and Xiong, Yongping and Zhang, Bo and others},
  booktitle={Proceedings of the IEEE/CVF Conference on Computer Vision and Pattern Recognition},
  pages={13691--13701},
  year={2025}
}

@inproceedings{fu2025video,
  title={Video-mme: The first-ever comprehensive evaluation benchmark of multi-modal llms in video analysis},
  author={Fu, Chaoyou and Dai, Yuhan and Luo, Yongdong and Li, Lei and Ren, Shuhuai and Zhang, Renrui and Wang, Zihan and Zhou, Chenyu and Shen, Yunhang and Zhang, Mengdan and others},
  booktitle={Proceedings of the IEEE/CVF conference on computer vision and pattern recognition},
  pages={24108--24118},
  year={2025}
}

@article{kumar2025mmau,
  title={Mmau-pro: A challenging and comprehensive benchmark for holistic evaluation of audio general intelligence},
  author={Kumar, Sonal and Sedl{\'a}{\v{c}}ek, {\v{S}}imon and Lokegaonkar, Vaibhavi and L{\'o}pez, Fernando and Yu, Wenyi and Anand, Nishit and Ryu, Hyeonggon and Chen, Lichang and Pli{\v{c}}ka, Maxim and Hlav{\'a}{\v{c}}ek, Miroslav and others},
  journal={arXiv preprint arXiv:2508.13992},
  year={2025}
}

@article{sakshi2024mmau,
  title={Mmau: A massive multi-task audio understanding and reasoning benchmark},
  author={Sakshi, Sakshi and Tyagi, Utkarsh and Kumar, Sonal and Seth, Ashish and Selvakumar, Ramaneswaran and Nieto, Oriol and Duraiswami, Ramani and Ghosh, Sreyan and Manocha, Dinesh},
  journal={arXiv preprint arXiv:2410.19168},
  year={2024}
}

@article{wang2025mmsu,
  title={Mmsu: A massive multi-task spoken language understanding and reasoning benchmark},
  author={Wang, Dingdong and Wu, Jincenzi and Li, Junan and Yang, Dongchao and Chen, Xueyuan and Zhang, Tianhua and Meng, Helen},
  journal={arXiv preprint arXiv:2506.04779},
  year={2025}
}

@article{ma2025mmar,
  title={Mmar: A challenging benchmark for deep reasoning in speech, audio, music, and their mix},
  author={Ma, Ziyang and Ma, Yinghao and Zhu, Yanqiao and Yang, Chen and Chao, Yi-Wen and Xu, Ruiyang and Chen, Wenxi and Chen, Yuanzhe and Chen, Zhuo and Cong, Jian and others},
  journal={arXiv preprint arXiv:2505.13032},
  year={2025}
}

@article{yang2025humanomniv2,
  title={Humanomniv2: From understanding to omni-modal reasoning with context},
  author={Yang, Qize and Yao, Shimin and Chen, Weixuan and Fu, Shenghao and Bai, Detao and Zhao, Jiaxing and Sun, Boyuan and Yin, Bowen and Wei, Xihan and Zhou, Jingren},
  journal={arXiv preprint arXiv:2506.21277},
  year={2025}
}

@article{li2025omnivideobench,
  title={Omnivideobench: Towards audio-visual understanding evaluation for omni mllms},
  author={Li, Caorui and Chen, Yu and Ji, Yiyan and Xu, Jin and Cui, Zhenyu and Li, Shihao and Zhang, Yuanxing and Wang, Wentao and Song, Zhenghao and Zhang, Dingling and others},
  journal={arXiv preprint arXiv:2510.10689},
  year={2025}
}

@article{zhou2025daily,
  title={Daily-omni: Towards audio-visual reasoning with temporal alignment across modalities},
  author={Zhou, Ziwei and Wang, Rui and Wu, Zuxuan},
  journal={arXiv preprint arXiv:2505.17862},
  year={2025}
}

@inproceedings{zhou2018towards,
  title={Towards automatic learning of procedures from web instructional videos},
  author={Zhou, Luowei and Xu, Chenliang and Corso, Jason},
  booktitle={Proceedings of the AAAI conference on artificial intelligence},
  volume={32},
  year={2018}
}

@misc{Farré2024FineVideo,
  title={FineVideo},
  author={Farré, Miquel and Marafioti, Andi and Tunstall, Lewis and Von Werra, Leandro and Wolf, Thomas},
  year={2024},
  howpublished={\url{https://huggingface.co/datasets/HuggingFaceFV/finevideo}},
}

@article{zhang2024llava,
  title={Llava-video: Video instruction tuning with synthetic data},
  author={Zhang, Yuanhan and Wu, Jinming and Li, Wei and Li, Bo and Ma, Zejun and Liu, Ziwei and Li, Chunyuan},
  journal={arXiv preprint arXiv:2410.02713},
  year={2024}
}

@misc{bai2025qwen25vltechnicalreport,
      title={Qwen2.5-VL Technical Report}, 
      author={Shuai Bai and Keqin Chen and Xuejing Liu and Jialin Wang and Wenbin Ge and Sibo Song and Kai Dang and Peng Wang and Shijie Wang and Jun Tang and Humen Zhong and Yuanzhi Zhu and Mingkun Yang and Zhaohai Li and Jianqiang Wan and Pengfei Wang and Wei Ding and Zheren Fu and Yiheng Xu and Jiabo Ye and Xi Zhang and Tianbao Xie and Zesen Cheng and Hang Zhang and Zhibo Yang and Haiyang Xu and Junyang Lin},
      year={2025},
      eprint={2502.13923},
      archivePrefix={arXiv},
      primaryClass={cs.CV},
      url={https://arxiv.org/abs/2502.13923}, 
}

@article{zhong2025omni,
  title={Omni-r1: Reinforcement learning for omnimodal reasoning via two-system collaboration},
  author={Zhong, Hao and Zhu, Muzhi and Du, Zongze and Huang, Zheng and Zhao, Canyu and Liu, Mingyu and Wang, Wen and Chen, Hao and Shen, Chunhua},
  journal={arXiv preprint arXiv:2505.20256},
  year={2025}
}

@article{feng2025video,
  title={Video-r1: Reinforcing video reasoning in mllms},
  author={Feng, Kaituo and Gong, Kaixiong and Li, Bohao and Guo, Zonghao and Wang, Yibing and Peng, Tianshuo and Wu, Junfei and Zhang, Xiaoying and Wang, Benyou and Yue, Xiangyu},
  journal={arXiv preprint arXiv:2503.21776},
  year={2025}
}

@article{shao2024deepseekmath,
  title={Deepseekmath: Pushing the limits of mathematical reasoning in open language models},
  author={Shao, Zhihong and Wang, Peiyi and Zhu, Qihao and Xu, Runxin and Song, Junxiao and Bi, Xiao and Zhang, Haowei and Zhang, Mingchuan and Li, YK and Wu, Yang and others},
  journal={arXiv preprint arXiv:2402.03300},
  year={2024}
}

@article{wang2025reinforcement,
  title={Reinforcement learning for reasoning in large language models with one training example},
  author={Wang, Yiping and Yang, Qing and Zeng, Zhiyuan and Ren, Liliang and Liu, Liyuan and Peng, Baolin and Cheng, Hao and He, Xuehai and Wang, Kuan and Gao, Jianfeng and others},
  journal={arXiv preprint arXiv:2504.20571},
  year={2025}
}

@inproceedings{sheng2025hybridflow,
  title={Hybridflow: A flexible and efficient rlhf framework},
  author={Sheng, Guangming and Zhang, Chi and Ye, Zilingfeng and Wu, Xibin and Zhang, Wang and Zhang, Ru and Peng, Yanghua and Lin, Haibin and Wu, Chuan},
  booktitle={Proceedings of the Twentieth European Conference on Computer Systems},
  pages={1279--1297},
  year={2025}
}

\clearpage
\appendix
\section{Appendix}
\label{sec:appendix}
\setcounter{section}{0}
\setcounter{subsection}{0}
\setcounter{subsubsection}{0}
\setcounter{secnumdepth}{2}
\renewcommand{\thesection}{A.\arabic{section}}
\makeatletter
\renewcommand\section{\@startsection{section}{1}{\z@}%
                       {-18\p@ \@plus -4\p@ \@minus -4\p@}%
                       {8\p@ \@plus 4\p@ \@minus 4\p@}%
                       {\normalfont\normalsize\bfseries\boldmath
                        \rightskip=\z@ \@plus 8em\pretolerance=10000 }}
\renewcommand\subsection{\@startsection{subsection}{2}{\z@}%
                       {-18\p@ \@plus -4\p@ \@minus -4\p@}%
                       {8\p@ \@plus 4\p@ \@minus 4\p@}%
                       {\normalfont\normalsize\bfseries\boldmath
                        \rightskip=\z@ \@plus 8em\pretolerance=10000 }}
\renewcommand\subsubsection{\@startsection{subsubsection}{3}{\z@}%
                       {-18\p@ \@plus -4\p@ \@minus -4\p@}%
                       {-0.5em \@plus -0.22em \@minus -0.1em}%
                       {\normalfont\normalsize\bfseries\boldmath}}
\providecommand*{\theHsection}{\thesection}
\providecommand*{\theHsubsection}{\thesubsection}
\providecommand*{\theHsubsubsection}{\thesubsubsection}
\providecommand*{\theHfigure}{\thefigure}
\providecommand*{\theHtable}{\thetable}
\providecommand*{\theHequation}{\theequation}
\providecommand*{\theHalgorithm}{\thealgorithm}
\renewcommand{\theHsection}{appendix.\thesection}
\renewcommand{\theHsubsection}{appendix.\thesubsection}
\renewcommand{\theHsubsubsection}{appendix.\thesubsubsection}
\renewcommand{\theHfigure}{appendix.figure.\arabic{figure}}
\renewcommand{\theHtable}{appendix.table.\arabic{table}}
\renewcommand{\theHequation}{appendix.equation.\arabic{equation}}
\renewcommand{\theHalgorithm}{appendix.algorithm.\arabic{algorithm}}
\makeatother
\noindent \textbf{Overview.} The appendix is organized as follows:

\begingroup
\footnotesize

\newcommand{\ovlink}[2]{%
    \hyperref[#1]{\textcolor{eccvblue}{\textbf{Section~\ref*{#1}:}}\ \textcolor{black}{\textbf{#2}}}%
}

\newcommand{\ovdesc}[1]{%
    \begin{itemize}[label={}, leftmargin=1.45em, topsep=0pt, itemsep=0pt, parsep=0pt, partopsep=0pt]
        \item #1
    \end{itemize}%
}

\newcommand{\ovitem}[2]{%
    \item \ovlink{#1}{#2}%
}

\newcommand{\ovleaf}[3]{%
    \item \ovlink{#1}{#2}%
    \ovdesc{#3}%
}

\begin{itemize}[label=\footnotesize$\blacktriangleright$, leftmargin=1.5em, topsep=2pt, itemsep=1pt, parsep=0pt, partopsep=0pt]

    \ovleaf{sec:unijigsaw_formulation}{Uni-Modal Jigsaw Formulation}
    {Defines the VideoJigsaw and AudioJigsaw variants used as uni-modal references.}

    \ovitem{sec:additional_implementation}{Additional Implementation Details}
    \begin{itemize}[label=\textbullet, leftmargin=1.25em, topsep=1pt, itemsep=1pt, parsep=0pt, partopsep=0pt]
        \ovleaf{sec:supp_data_filtering}{Details on Data Filtering Pipeline}
        {Describes both the heuristic signal-based filtering algorithm and the MLLM-based semantic screening configurations.}

        \ovleaf{sec:supp_training}{Details on Training}
        {Describes training data preparation, preprocessing, and hyperparameter settings for OmniJigsaw post-training.}

        \ovleaf{sec:supp_evaluation}{Details on Evaluation}
        {Describes the benchmark suite, inference input preprocessing, and evaluation parameter settings.}
    \end{itemize}

    \ovitem{sec:more_results}{More Results}
    \begin{itemize}[label=\textbullet, leftmargin=1.25em, topsep=1pt, itemsep=1pt, parsep=0pt, partopsep=0pt]
        \ovleaf{sec:supp_data_cases}{Cases of Semantic-based Data Filtering}
        {Provides representative semantically rejected examples that illustrate why additional screening is necessary.}

        \ovleaf{sec:supp_fine_grained_results}{Fine-grained Evaluation across Sub-capabilities}
        {Reports sub-capability evaluation results on representative video, audio, and omni-modal benchmarks.}

        \ovleaf{sec:supp_qual_examples}{Qualitative Examples}
        {Presents qualitative comparisons between the baseline and its OmniJigsaw (CMM)-post-trained variant.}
    \end{itemize}

    \ovleaf{sec:limitations_future_work}{Limitations and Future Work}
    {Discusses the current limitations of OmniJigsaw and possible future directions.}

    \ovitem{sec:prompts}{Prompts}
    \begin{itemize}[label=\textbullet, leftmargin=1.25em, topsep=1pt, itemsep=1pt, parsep=0pt, partopsep=0pt]
        \ovleaf{sec:semantic_screening_prompt}{Semantic Screening Prompt}
        {Presents the prompt used for MLLM-based semantic screening.}

        \ovleaf{sec:training_prompts}{Training Prompts}
        {Lists the prompts used for OmniJigsaw under different modality orchestration strategies.}

        \ovleaf{sec:evaluation_prompts}{Evaluation Prompts}
        {Provides the prompt templates used for benchmark evaluation.}
    \end{itemize}

\end{itemize}
\endgroup

\vspace{0.15em}

\section{Uni-Modal Jigsaw Formulation}
\label{sec:unijigsaw_formulation}

To comprehensively investigate the modality-specific enhancements yielded by OmniJigsaw, we establish two uni-modal references: VideoJigsaw and AudioJigsaw. Consistent with the notation defined in the OmniJigsaw Formulation section, we denote an omni-modal sample as $\mathcal{X} = (\mathcal{V}, \mathcal{A})$, which is segmented into $N$ non-overlapping clips $\mathcal{S} = [s_1, s_2, \dots, s_N]$, where each clip $s_i = (v_i, a_i)$ contains synchronized visual and acoustic information. Let $\pi$ be a random permutation sampled from the set of all possible bijective mappings of the index set $\{1, \dots, N\}$ onto itself. The goal of these uni-modal tasks is to recover the ground truth chronological sequence $\mathbf{y} = [\pi(1), \pi(2), \dots, \pi(N)]$ by observing only a single modality.

VideoJigsaw requires the model $\mathcal{M}_\theta$ to reconstruct the temporal order relying solely on visual clips from the shuffled sequence, explicitly excluding any acoustic signals. This is formalized as:
\begin{equation}
    \hat{\mathbf{y}}_v = \mathcal{M}_\theta(\tilde{\mathcal{V}}; \mathcal{I}_{v}), \quad \text{where } \tilde{\mathcal{V}} = [v_{\pi^{-1}(1)}, v_{\pi^{-1}(2)}, \dots, v_{\pi^{-1}(N)}].
\end{equation}
The input $\tilde{\mathcal{V}}$ consists of $N$ visual clips, where the $j$-th disordered clip corresponds to the $\pi^{-1}(j)$-th clip in the original chronological sequence.

Similarly, AudioJigsaw requires the model $\mathcal{M}_\theta$ to reorder shuffled audio clips without any visual assistance. The formulation is correspondingly defined as:
\begin{equation}
    \hat{\mathbf{y}}_a = \mathcal{M}_\theta(\tilde{\mathcal{A}}; \mathcal{I}_{a}), \quad \text{where } \tilde{\mathcal{A}} = [a_{\pi^{-1}(1)}, a_{\pi^{-1}(2)}, \dots, a_{\pi^{-1}(N)}].
\end{equation}

By decoupling the omni-modal inputs into these modality-specific variants, we can comprehensively assess and quantify the respective gains of each modality achieved through our proposed modality orchestration strategies.

\section{Additional Implementation Details}
\label{sec:additional_implementation}

\subsection{Details on Data Filtering Pipeline}
\label{sec:supp_data_filtering}

\subsubsection{Algorithmic Implementation of Heuristic Filtering}
\label{sec:details_of_stage_1}

To guarantee the fundamental solvability of the temporal puzzles, we initially develop a heuristic filtering pipeline to prune ill-posed samples that lack modal integrity or irreversible transitions. The complete algorithmic workflow is systematically formalized in Algorithm~\ref{alg:heuristic_filtering}.

\begin{algorithm}[t!]
\caption{Signal-based Heuristic Filtering Pipeline}
\label{alg:heuristic_filtering}
\begin{algorithmic}[1]
\Require Raw video dataset $\mathcal{V}_{raw}$.
\Ensure Filtered and standardized video dataset $\mathcal{V}_{filtered}$.
\State \textbf{Hyperparameters:} Duration limit $D_{max} = 200$\,s (adjustable based on computational resources); Visual sample interval $\Delta t = 1.0$\,s; MAD threshold $\tau_{v} = 5.0$; Max static ratio $R_{v\_max} = 0.70$; Target sample rate $SR = 16$\,kHz; RMS silence threshold $\tau_{rms} = -40$\,dB; Max silence ratio $R_{s\_max} = 0.70$; Min SF variance $\tau_{sf} = 0.5$; VAD ratio bounds $[R_{vad\_min}, R_{vad\_max}] = [0.30, 0.80]$.
\State $\mathcal{V}_{filtered} \gets \emptyset$
\For{each video $V \in \mathcal{V}_{raw}$}
    \State Extract duration $D$ and stream metadata via \texttt{ffprobe}.
    \If{$D > D_{max}$ \textbf{or} $V$ lacks audio/video streams}
        \State \textbf{continue} \Comment{Discard invalid or overly long videos}
    \EndIf

    \State \textit{\% Stage 1: Visual Dynamism Assessment}
    \State Uniformly sample frame sequence $\{I_1, I_2, \dots, I_N\}$ at interval $\Delta t$.
    \State $C_{static} \gets 0$
    \For{$i = 2$ \textbf{to} $N$}
        \State $\hat{I}_i \gets \text{Grayscale}(\text{Resize}(I_i, 64 \times 64))$
        \State $\text{MAD}_i \gets \frac{1}{H \times W} \sum |\hat{I}_{i} - \hat{I}_{i-1}|$
        \If{$\text{MAD}_i < \tau_{v}$}
            \State $C_{static} \gets C_{static} + 1$
        \EndIf
    \EndFor
    \If{$C_{static} / (N-1) > R_{v\_max}$} 
        \State \textbf{continue} \Comment{Discard visually static videos}
    \EndIf

    \State \textit{\% Stage 2: Audio Quality and Information Density}
    \State Extract audio $A_{16k} \gets \text{Resample}(V_{audio}, SR)$
    
    \State $\text{RMS}_{dB} \gets 20 \log_{10}(\text{RMS\_Energy}(A_{16k}) / \max(\text{RMS\_Energy}))$
    \State $\text{Ratio}_{silence} \gets \text{Count}(\text{RMS}_{dB} < \tau_{rms}) / \text{Length}(\text{RMS}_{dB})$
    \If{$\text{Ratio}_{silence} > R_{s\_max}$} \textbf{continue} \EndIf
    
    \State $\text{Var}_{SF} \gets \text{Variance}(\text{Onset\_Strength}(A_{16k}))$
    \If{$\text{Var}_{SF} < \tau_{sf}$} \textbf{continue} \EndIf  \Comment{Discard monotone noise}

    \State $\mathcal{T}_{speech} \gets \text{Silero\_VAD}(A_{16k})$
    \State $\text{Ratio}_{speech} \gets \sum (\text{durations in } \mathcal{T}_{speech}) / D$
    \If{$\text{Ratio}_{speech} \notin [R_{vad\_min}, R_{vad\_max}]$} \textbf{continue} \EndIf

    \State \textit{\% Stage 3: Standardization}
    \State Standardize $V$ to H.264/AAC in \texttt{.mp4} format.
    \State $\mathcal{V}_{filtered} \gets \mathcal{V}_{filtered} \cup \{V\}$
\EndFor
\State \textbf{return} $\mathcal{V}_{filtered}$
\end{algorithmic}
\end{algorithm}

First, we parse the metadata of each raw video. To ensure omni-modal integrity, samples lacking either a valid visual or audio stream are strictly discarded. Furthermore, taking computational tractability into account, samples exceeding a maximum duration of $D_{max}$ can be further excluded. $D_{max}$ is highly flexible and can be dynamically adjusted according to the available computational resources (we set $D_{max} = 200$ seconds).

Subsequently, we evaluate visual dynamism. To optimize processing efficiency, we uniformly sample frames at an interval of $1.0$ second. Each sampled frame is spatially downsampled to $64 \times 64$ pixels and converted to grayscale. We then compute the Mean Absolute Difference (MAD) between adjacent sampled frames. A frame transition is classified as static if the MAD falls below a strict threshold of $\tau_{v} = 5.0$ (on a $0$--$255$ intensity scale). Videos exhibiting a static frame ratio exceeding $70\%$ are pruned, thereby eliminating frozen screens or visually monotonous content. 

For the audio stream, we extract a mono-channel signal resampled to $16$\,kHz. The audio evaluation is tripartite: (1) Silence Removal: We compute the Root Mean Square (RMS) amplitude and convert it to a Decibel (dB) scale relative to the maximum amplitude. Audio segments below $-40$\,dB are flagged as silence. Videos with a silence ratio greater than $70\%$ are removed. (2) Dynamics Assessment: To filter out constant or monotone background noise, we calculate the onset strength envelope, representing the Spectral Flux (SF). A minimum SF variance of $0.5$ is mandated, thereby excluding signals that lack sufficient acoustic dynamics. (3) Information Density Check: We employ a pretrained Silero Voice Activity Detection (VAD) model to locate human speech. We require the total speech ratio to constitute between $30\%$ and $80\%$ of the video length. This bounded ratio deliberately ensures that the video contains rich instructional speech without overwhelmingly dominating the ambient acoustic events. 

Finally, all surviving candidates are standardized into the MP4 container with H.264 video and AAC audio encodings.

\subsubsection{Inference Configurations for Semantic Screening}
\label{sec:details_of_stage_2}

For MLLM-based semantic screening, we deploy Qwen2.5-VL-7B-Instruct~\cite{bai2025qwen25vltechnicalreport} as the semantic assessor to guarantee temporal irreversibility and distinct state transitions within the candidate videos. Unlike heuristic filters that operate on low-level signals, this stage focuses on high-level causal progression and logical determinism essential for OmniJigsaw. To achieve a comprehensive understanding of long-range temporal dynamics, we configure the model to process $200$ uniformly downsampled frames per video, with a maximum spatial resolution of $100,352$ pixels per frame. 

To ensure assessment reliability, the model's inference parameters are meticulously tuned, as summarized in Table~\ref{tab:inference_config}. Specifically, we adopt a greedy decoding strategy (temperature $\tau = 0$) and a repetition penalty of $1.05$ to suppress degenerative loops during long-form generation. We formally instruct the model to adhere to a Chain-of-Thought (CoT) protocol, wherein it must encapsulate its qualitative analysis, including evaluation criteria such as causal progression, visual evolution, and temporal markers, within \texttt{<think>...</think>} tags prior to emitting a definitive \texttt{YES/NO} decision within \texttt{<answer>...</answer>} tags. A maximum generation limit of $2,048$ tokens is allocated to accommodate exhaustive reasoning. This structured reasoning process ensures that only videos with unambiguous chronological flow and irreversible state changes are retained, effectively excluding repetitive actions or visually ambiguous content that would otherwise make the jigsaw puzzle ill-posed.

\begin{table}[t!]
\centering
\caption{\textbf{Inference configurations for semantic screening via Qwen2.5-VL-7B-Instruct~\cite{bai2025qwen25vltechnicalreport}.}}
\label{tab:inference_config}
\fontfamily{ptm}\selectfont
\setlength{\tabcolsep}{8pt}
\begin{tabular}{@{}llc@{}}
\toprule
\textbf{Category} & \textbf{Parameter} & \textbf{Value} \\ \midrule
\multirow{2}{*}{Visual Input} & Maximum Sampled Frames & 200 \\
 & Maximum Resolution (pixels) & 100,352 \\ \midrule
\multirow{4}{*}{Sampling Params} & Temperature ($\tau$) & 0 (Greedy Decoding) \\
 & Top-$p$ & 1.0 \\
 & Top-$k$ & -1 (Disabled) \\
 & Repetition Penalty & 1.05 \\ \midrule
\multirow{2}{*}{Generation Specs} & Maximum New Tokens & 2,048 \\
 & Context Length Limit & 32,768 \\
\bottomrule
\end{tabular}
\end{table}

\subsection{Details on Training}
\label{sec:supp_training}

\subsubsection{Data Preparation}
\label{sec:data_preparation}
To construct our training dataset, denoted as OmniJigsaw-8K, we aggregate raw videos from multiple high-quality sources to ensure diversity in temporal logic: YouCook2~\cite{zhou2018towards} is utilized for its procedural instructional causal chains, FineVideo~\cite{Farré2024FineVideo} for coherent narrative flows, and the NextQA subset of LLaVA-Video-178K~\cite{zhang2024llava} for capturing dynamic physical actions. These raw data are subsequently curated through our two-stage data filtering pipeline to eliminate ill-posed instances. The progressive filtering statistics throughout this curation process are detailed in Table~\ref{tab:data_stats}. Ultimately, we obtain $8,220$ high-fidelity samples, which are randomly partitioned into a training set of $8,156$ instances and a validation set of $64$ instances.

\begin{table}[t!]
    \centering
    \caption{\textbf{Data filtering statistics for OmniJigsaw-8K curation.} The pipeline progressively distills massive unannotated omni-modal data into high-fidelity samples suitable for the jigsaw proxy task.}
    \label{tab:data_stats}
    \fontfamily{ptm}\selectfont
    \renewcommand{\arraystretch}{1.2}
    \setlength{\tabcolsep}{2pt}
    \footnotesize
    \begin{tabular}{@{}lccc@{}}
        \toprule
        \textbf{Source Dataset} & \textbf{Raw Samples} & \textbf{After Stage 1} & \textbf{After Stage 2} \\
        \midrule
        YouCook2~\cite{zhou2018towards} & 2,000 & 327 & 327 \\
        FineVideo~\cite{Farré2024FineVideo} & 43,751 & 7,737 & 6,986 \\
        LLaVA-Video-178K~\cite{zhang2024llava} (NextQA) & 3,868 & 982 & 907 \\
        \midrule
        \rowcolor{gray!15} \textbf{Total} & \textbf{49,619} & \textbf{9,046} & \textbf{8,220} \\
        \bottomrule
    \end{tabular}
\end{table}

\subsubsection{Data Preprocessing}
\label{sec:supp_data_preprocessing}
Consistent with the OmniJigsaw formulation, the data preprocessing pipeline is meticulously designed to construct robust temporal reordering puzzles. Given an original video sample $\mathcal{X} = (\mathcal{V}, \mathcal{A})$, the sequence is uniformly partitioned into $N=6$ non-overlapping clips. To prevent trivial solutions derived from low-level boundary continuity (\eg, matching identical consecutive frames across clip boundaries), the trimming operator $\mathcal{T}_{trim}$ explicitly discards $5\%$ of the temporal duration from both the beginning and the end of each clip. Crucially, to guarantee exact multimodal synchronization and tensor uniformity, $\mathcal{T}_{trim}$ is applied to both visual and acoustic streams, forcing all clips to align with the duration of the shortest trimmed clip.

Regarding intra-clip preprocessing, the visual temporal downsampling operator $\mathcal{D}_T$ extracts frames uniformly via linearly spaced sampling at a default target rate of $2.0$\,FPS. To ensure computational tractability while preserving dynamic integrity, the sampled frame count per clip is strictly bounded between $2$ and $12$ frames. Subsequently, visual frames undergo an aspect-ratio-preserving spatial rescaling. Specifically, the target resolution is dynamically calculated to ensure that the total pixel count remains within $100,352$ while preserving the original aspect ratio. Concurrently, the synchronized audio waveform of each trimmed clip is resampled to $16$\,kHz without further truncation.

Based on these fundamental preprocessing protocols, the strategy-specific modality orchestrations, governed by the function $\Phi(\cdot)$, are implemented as follows:

\begin{itemize}[label=\footnotesize$\blacktriangleright$, leftmargin=1.5em]
    \item \textbf{OmniJigsaw (JMI):} As an identity mapping, $\Phi_{jmi}$ retains the complete synchronized visual and acoustic tensors $(\mathcal{D}_T(v_i), a_i)$ for each clip.
    
    \item \textbf{OmniJigsaw (SMS):} $\Phi_{sms}$ operates via a two-stage mechanism. In the first phase, a dominance analyzer (Qwen3-Omni-30B-A3B-Instruct~\cite{xu2025qwen3}) evaluates the global modality dominance $d \in \{V, A\}$ by ingesting a temporally compressed global visual context (downsampled at $1.0$\,FPS and capped at $80$ frames) alongside the full audio track. In the second phase, guided strictly by $d$, $\Phi_{sms}$ completely discards the unselected modality, ensuring that the jigsaw rollout is entirely uni-modal to effectively mitigate interference from the less informative stream.
    
    \item \textbf{OmniJigsaw (CMM):} Building upon the JMI pipeline, $\mathcal{M}_{cmm}$ executes fine-grained clip-level modality masking according to the modality selection vector $\mathbf{m}$ (using Qwen3-Omni-30B-A3B-Instruct~\cite{xu2025qwen3} as the modality selector). By replacing unselected modalities with $\mathbf{0}$, this strategy imposes a cross-modal information bottleneck that necessitates dynamic switching of attention between visual and acoustic cues.
\end{itemize}

\begin{table}[t!]
    \centering
    \caption{\textbf{Hyperparameter settings for OmniJigsaw GRPO post-training implemented via VeRL~\cite{sheng2025hybridflow}.}}
    \label{tab:hyperparams}
    \fontfamily{ptm}\selectfont
    \renewcommand{\arraystretch}{1.1}
    \setlength{\tabcolsep}{10pt}
    \footnotesize
    \begin{tabular}{lc}
        \toprule
        \textbf{Hyperparameter} & \textbf{Value} \\
        \midrule
        \rowcolor{gray!15} \multicolumn{2}{c}{\textit{Model} \& \textit{Infrastructure}} \\
        Base Policy Model & Qwen3-Omni-30B-A3B-Instruct~\cite{xu2025qwen3} \\
        Hardware Infrastructure & $8 \times$ NVIDIA H200 GPUs \\
        Frozen Components & Vision Tower, Audio Tower, Router \\
        \midrule
        \rowcolor{gray!15} \multicolumn{2}{c}{\textit{Training Dynamics (VeRL)}} \\
        Total Training Steps & 1,000 \\
        Global Batch Size & 8 \\
        PPO Mini Batch Size & 8 \\
        PPO Micro Batch Size (per GPU) & 4 \\
        Learning Rate & $1 \times 10^{-6}$ \\
        KL Penalty Coefficient & $1 \times 10^{-2}$ \\
        KL Estimator Type & \texttt{low\_var\_kl} \\
        \midrule
        \rowcolor{gray!15} \multicolumn{2}{c}{\textit{Prompt} \& \textit{Rollout Config}} \\
        Max Prompt Length & 8,192 \\
        Max Response Length & 2,048 \\
        Rollout Group Size & 8 \\
        Decoding Temperature & 0.9 \\
        Top-$k$ & 50 \\
        Top-$p$ & 1.0 \\
        Inference Repetition Penalty & 1.05 \\
        \midrule
        \rowcolor{gray!15} \multicolumn{2}{c}{\textit{Reward Configuration}} \\
        Positional Weight ($w_{pos}$) & 0.5 \\
        Continuity Weight ($w_{cont}$) & 0.5 \\
        Format Reward ($R_{fmt}$) & +0.2 \\
        Discount Factor $\lambda(\text{acc})$ & 1.0 (acc=1), 0.2 (acc=0) \\
        Repetition Penalty ($R_{rep}$) & -0.5 \\
        \bottomrule
    \end{tabular}
\end{table}

\subsubsection{Hyperparameter Settings}
\label{sec:supp_hyperparameter_settings}
We adopt Qwen3-Omni-30B-A3B-Instruct~\cite{xu2025qwen3} as the initial policy model and implement GRPO RL post-training under the proposed modality orchestration strategies. Our training pipeline is built upon the Volcano Engine Reinforcement Learning (VeRL)~\cite{sheng2025hybridflow} framework, and all experiments are conducted on a cluster of $8 \times$ NVIDIA H200 GPUs. Accounting for computational constraints and training efficiency, we freeze the vision tower, audio tower, and router across all strategies to prioritize reasoning alignment over perceptual feature learning. The training proceeds for $1,000$ steps with a global batch size of $8$. We employ on-policy learning with a learning rate of $1 \times 10^{-6}$ and a KL penalty coefficient of $1 \times 10^{-2}$ estimated via the \texttt{low\_var\_kl} method, which serves to avert catastrophic forgetting from potential overly aggressive updates. During the sampling process, we generate $8$ responses per prompt with a decoding temperature of $0.9$, a top-$k$ of $50$, and a top-$p$ of $1.0$. Furthermore, a repetition penalty of $1.05$ is applied to mitigate degenerative looping and linguistic redundancy, which are typical challenges in complex multimodal reasoning. Regarding the reward configuration, the composite function $R_{tot}$ is parameterized with $w_{pos}=0.5$ and $w_{cont}=0.5$. The accuracy-dependent discount $\lambda(\text{acc})$ is set to $1.0$ for perfect reordering and $0.2$ otherwise. Additionally, a format reward $R_{fmt}=+0.2$ is awarded for structural adherence, while a repetition penalty $R_{rep}=-0.5$ is triggered by $N$-gram ($N=20$) frequencies exceeding $3$. Detailed hyperparameter configurations are summarized in Table~\ref{tab:hyperparams}.

\subsection{Details on Evaluation}
\label{sec:supp_evaluation}

\subsubsection{Notes on Evaluation Benchmarks}
\label{sec:supp_benchmarks}

We evaluate OmniJigsaw across $15$ diverse benchmarks covering video, audio, and omni-modal reasoning. We primarily focus on the multiple-choice QA within these benchmarks and report the top-1 accuracy as the primary performance metric. The evaluated benchmarks and their corresponding evaluation files are summarized in Table~\ref{tab:evaluation_benchmarks}.

\begin{table}[t!]
    \centering
    \caption{\textbf{Detailed list of evaluation benchmarks and files.}}
    \label{tab:evaluation_benchmarks}
    \fontfamily{ptm}\selectfont
    \renewcommand{\arraystretch}{1.1}
    \setlength{\tabcolsep}{7pt}
    \footnotesize
    \begin{tabular}{@{}lll@{}}
        \toprule
        \textbf{Modality} & \textbf{Benchmark} & \textbf{Evaluation File} \\
        \midrule
        \multirow{14}{*}{Video} & AoTBench~\cite{xue2025seeing} & \texttt{data\_files/AoTBench\_QA.json} \\
        & TUNA-Bench~\cite{kong2025tuna} & \texttt{TUNA-MCQ.json} \\
        & TempCompass~\cite{liu2024tempcompass} & \texttt{multi-choice/test-00000-of-00001.parquet} \\
        & Video-TT~\cite{zhang2025towards} & \texttt{data/test-00000-of-00001.parquet} \\
        & Video-Holmes~\cite{cheng2025video} & \texttt{test\_Video-Holmes.json} \\
        & MLVU-Test~\cite{zhou2025mlvu} & \texttt{test\_multi\_choice\_tasks.json} \\
        & Video-MME~\cite{fu2025video} & \texttt{videomme/test-00000-of-00001.parquet} \\
        \cmidrule{2-3}
        & \multirow{7}{*}{MLVU~\cite{zhou2025mlvu}} & \texttt{MLVU/json/1\_plotQA.json} \\
        &  & \texttt{MLVU/json/2\_needle.json} \\
        &  & \texttt{MLVU/json/3\_ego.json} \\
        &  & \texttt{MLVU/json/4\_count.json} \\
        &  & \texttt{MLVU/json/5\_order.json} \\
        &  & \texttt{MLVU/json/6\_anomaly\_reco.json} \\
        &  & \texttt{MLVU/json/7\_topic\_reasoning.json} \\
        \midrule
        \multirow{4}{*}{Audio} & MMAU-Pro~\cite{kumar2025mmau} & \texttt{test.parquet} \\
        & MMAU-test-mini~\cite{sakshi2024mmau} & \texttt{mmau-test-mini.json} \\
        & MMSU~\cite{wang2025mmsu} & \texttt{question/mmsu.jsonl} \\
        & MMAR~\cite{ma2025mmar} & \texttt{MMAR-meta.json} \\
        \midrule
        \multirow{3}{*}{Omni-modal} & Daily-Omni~\cite{zhou2025daily} & \texttt{qa.json} \\
        & IntentBench~\cite{yang2025humanomniv2} & \texttt{qa.json} \\
        & OmniVideoBench~\cite{li2025omnivideobench} & \texttt{data.parquet} \\
        \bottomrule
    \end{tabular}
\end{table}

\begin{table}[t!]
    \centering
    \caption{\textbf{Evaluation sampling parameter configurations.} These settings are uniformly applied across all benchmark evaluations to ensure deterministic and reproducible outputs.}
    \label{tab:inference_params}
    \fontfamily{ptm}\selectfont
    \renewcommand{\arraystretch}{1.2}
    \setlength{\tabcolsep}{12pt}
    \footnotesize
    \begin{tabular}{@{}lc@{}}
        \toprule
        \textbf{Parameter} & \textbf{Value} \\
        \midrule
        Temperature ($\tau$) & 0 (Greedy Decoding) \\
        Top-$p$ & 1.0 \\
        Top-$k$ & -1 (Disabled) \\
        Max New Tokens & 2,048 \\
        Repetition Penalty & 1.05 \\
        \bottomrule
    \end{tabular}
\end{table}

\subsubsection{Inference Input Preprocessing}
\label{sec:supp_inference_preprocessing}

To ensure standardized evaluation, the inference input preprocessing protocols are dynamically instantiated based on the specific reasoning modes and modalities. The detailed video and audio processing logic for each evaluation mode is formalized as follows:

\begin{itemize}[label=\footnotesize$\blacktriangleright$, leftmargin=1.5em]
    \item \textbf{Video Reasoning:}
    \begin{itemize}[label=\textbullet, leftmargin=1.2em]
        \item \textbf{Infer w/ Audio:} 
        \begin{itemize}[label=\footnotesize$\circ$, leftmargin=1.2em]
            \item \textit{Video:} The visual stream is uniformly sampled to a maximum of $200$ frames. Each frame undergoes an aspect-ratio-preserving spatial rescaling to ensure that the total pixel count does not exceed $100,352$ pixels.
            
            \item \textit{Audio:} The acoustic stream is resampled to $16$\,kHz with a maximum duration limit of $600$ seconds. An adaptive audio extraction strategy is applied based on the temporal span of the sampled video frames: 
            \begin{enumerate}[label=(\alph*), leftmargin=1.8em]
                \item Continuous Extraction: If the temporal span between the first and last sampled frames is within $600$ seconds, a continuous audio segment centered on the midpoint of the frame sequence is extracted.
                \item Frame-Anchored Extraction: Conversely, if the temporal span exceeds $600$ seconds, the maximum duration limit of $600$ seconds is equally divided by the total number of sampled frames. Short audio chunks centered at the exact timestamp of each visual frame are extracted, zero-padded if necessary, and sequentially concatenated.
            \end{enumerate}
        \end{itemize}
        
        \item \textbf{Infer w/o Audio:} The visual stream is processed identically to the Infer w/ Audio mode.
    \end{itemize}

    \item \textbf{Audio Reasoning:} The raw audio waveform is resampled to $16$\,kHz with a maximum duration limit of $3,600$ seconds. If the audio length exceeds this threshold, a central cropping operation is performed to strictly retain the middle $3,600$ seconds. Missing audio streams are universally padded with zero arrays to maintain inference stability.

    \item \textbf{Omni-Modal Collaborative Reasoning:} Inputs are processed identically to the Infer w/ Audio mode in the Video Reasoning setting.
\end{itemize}

\subsubsection{Inference Parameter Settings}
\label{sec:supp_inference_params}
To ensure the determinism and reproducibility of the model's reasoning processes, all benchmark evaluations are conducted using a greedy decoding strategy. We apply a repetition penalty of $1.05$ to mitigate potential degenerative looping. The maximum number of newly generated tokens is set to $2,048$ to accommodate the extensive reasoning often required for complex video, audio, and omni-modal QA tasks. The detailed configurations for the evaluation sampling parameters are summarized in Table~\ref{tab:inference_params}.

\section{More Results}
\label{sec:more_results}

\subsection{Cases of Semantic-based Data Filtering}
\label{sec:supp_data_cases}

To further demonstrate the efficacy of our two-stage data filtering pipeline, \afterpage{
\clearpage
\begin{figure}[p]
    \centering
    \includegraphics[width=1.0\textwidth, height=0.96\textheight, keepaspectratio]{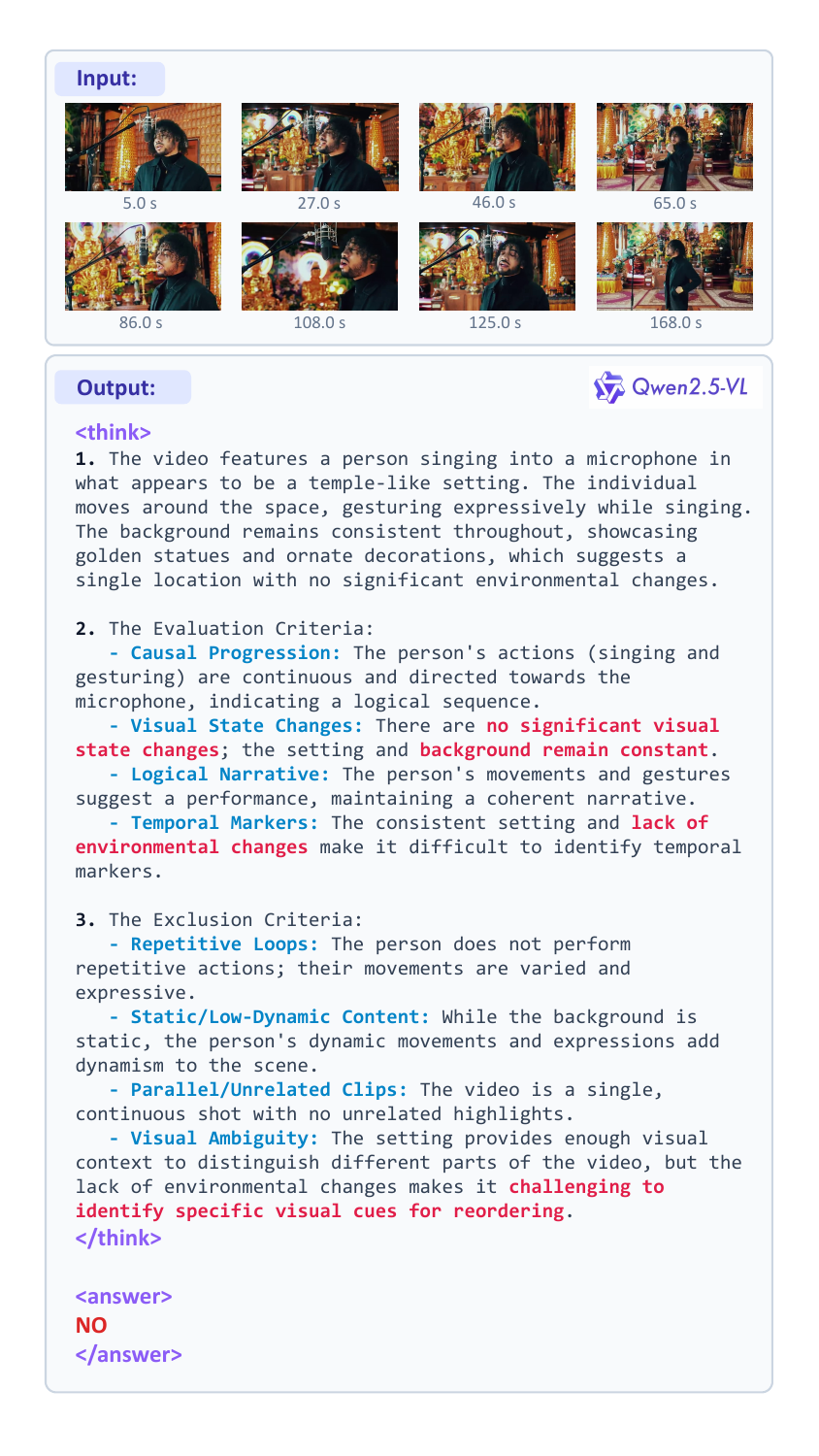} 
    \vspace*{-4mm}
    \caption{\textbf{Case 1: Indistinct State Changes.}}
    \label{fig:rejection_case_1}
\end{figure}
\clearpage
\begin{figure}[t!]
    \centering
    \includegraphics[width=1.0\textwidth, height=0.96\textheight, keepaspectratio]{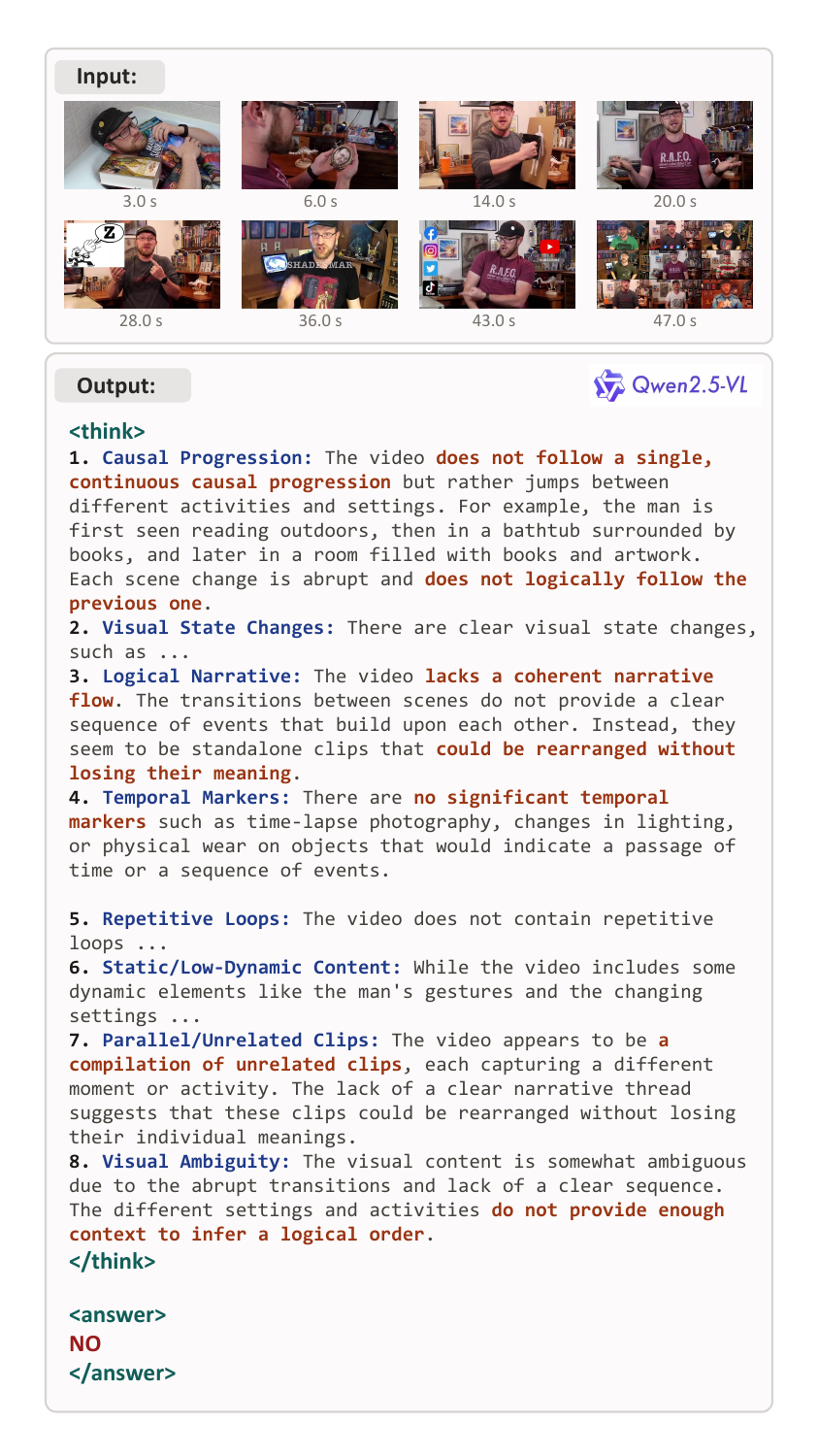} 
    \vspace*{-4mm}
    \caption{\textbf{Case 2: Disjointed Narrative.}}
    \label{fig:rejection_case_2}
\end{figure}
\clearpage
}we provide representative qualitative examples of videos that pass initial signal-based heuristic filtering but are subsequently rejected during the semantic-based reasoning screening.

\subsubsection{Case 1: Indistinct State Changes}
\label{sec:case_1}
As illustrated in Figure~\ref{fig:rejection_case_1}, the video shows a person singing and gesturing in a temple-like indoor scene with consistent background elements (\eg, microphone, statues, and decorations) throughout its duration. Although visually appealing and continuously dynamic, it is not suitable for the jigsaw proxy task. As indicated by the MLLM (Qwen2.5-VL-7B-Instruct~\cite{bai2025qwen25vltechnicalreport}) output, this sample exhibits coherent performance dynamics but lacks sufficiently distinct visual states, clear environmental transitions, and reliable temporal markers for reordering. Consequently, while the video is not static in a low-level signal sense, it still lacks a directionally unambiguous chronological progression that would allow shuffled clips to be restored. This case illustrates the necessity of our semantic-based reasoning screening, which further removes videos with mild temporal variation but insufficient temporal-causal structure, ensuring that the retained samples exhibit strong and recoverable temporal reordering cues.

\subsubsection{Case 2: Disjointed Narrative}
\label{sec:case_2}
Figure~\ref{fig:rejection_case_2} shows another semantically rejected sample with a disjointed narrative. Although this video contains noticeable scene changes, it remains unsuitable for the jigsaw proxy task. The video jumps across visually distinct yet weakly related scenes (\eg, sitting in a bathtub with books and appearing in an indoor room filled with books and artwork), rather than forming a single causal progression. The model explicitly identifies the lack of a coherent narrative flow, the absence of reliable temporal markers, and the fact that the clips could be rearranged without losing their meaning, and therefore returns a final \texttt{NO} decision. This example highlights that visual diversity alone does not imply suitability for the jigsaw proxy task. It further demonstrates the necessity of our semantic-based reasoning screening, which removes videos with apparent state changes but without directionally unambiguous chronological progression, thereby curating high-quality samples characterized by robust temporal-causal structures that are well suited for the jigsaw proxy task.

\subsection{Fine-grained Evaluation across Sub-capabilities}
\label{sec:supp_fine_grained_results}

The sub-capability evaluation results of OmniJigsaw on Video-MME~\cite{fu2025video} (video reasoning), MMAU-test-mini~\cite{sakshi2024mmau} (audio reasoning), and OmniVideoBench~\cite{li2025omnivideobench} (omni-modal collaborative reasoning) are systematically reported in Tables~\ref{tab:video_mme_sub},~\ref{tab:mmau_sub}, and~\ref{tab:omnivideobench_sub}, respectively. Overall, the three modality orchestration strategies exhibit distinct strengths across benchmarks and sub-capabilities. In particular, they show relatively consistent improvements on several sub-capabilities in Video-MME and MMAU-test-mini, while the gains on OmniVideoBench are more heterogeneous, reflecting the greater difficulty of omni-modal collaborative reasoning.

\begin{figure*}[tp]
    \centering
    \includegraphics[width=\textwidth]{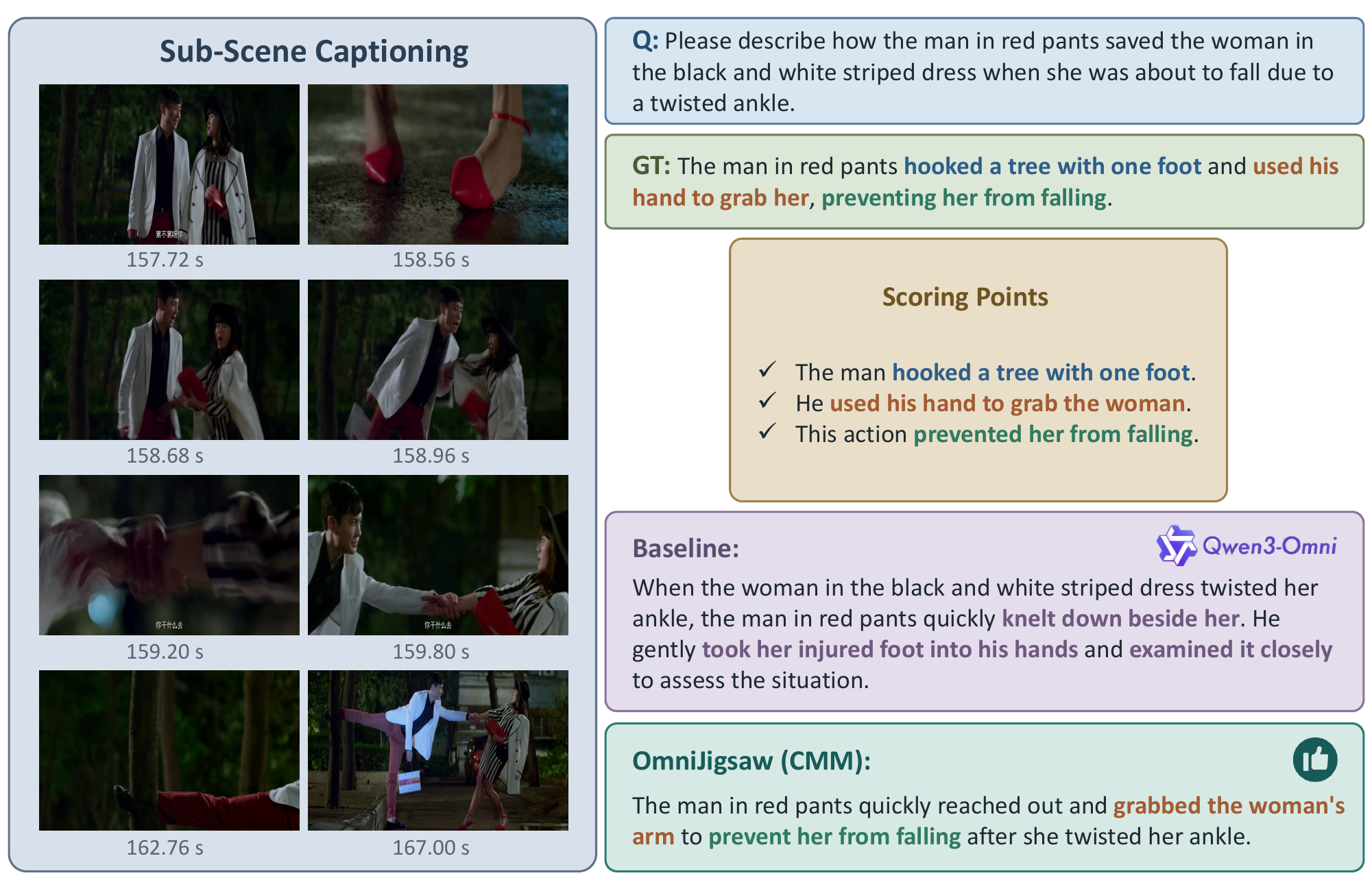}
    \caption{\textbf{Qualitative example of Sub-Scene Captioning.} Comparison between the Qwen3-Omni-30B-A3B-Instruct~\cite{xu2025qwen3} baseline and its OmniJigsaw (CMM)-post-trained variant.}
    \label{fig:qual_subscene_captioning}

    \centering
    \includegraphics[width=\textwidth]{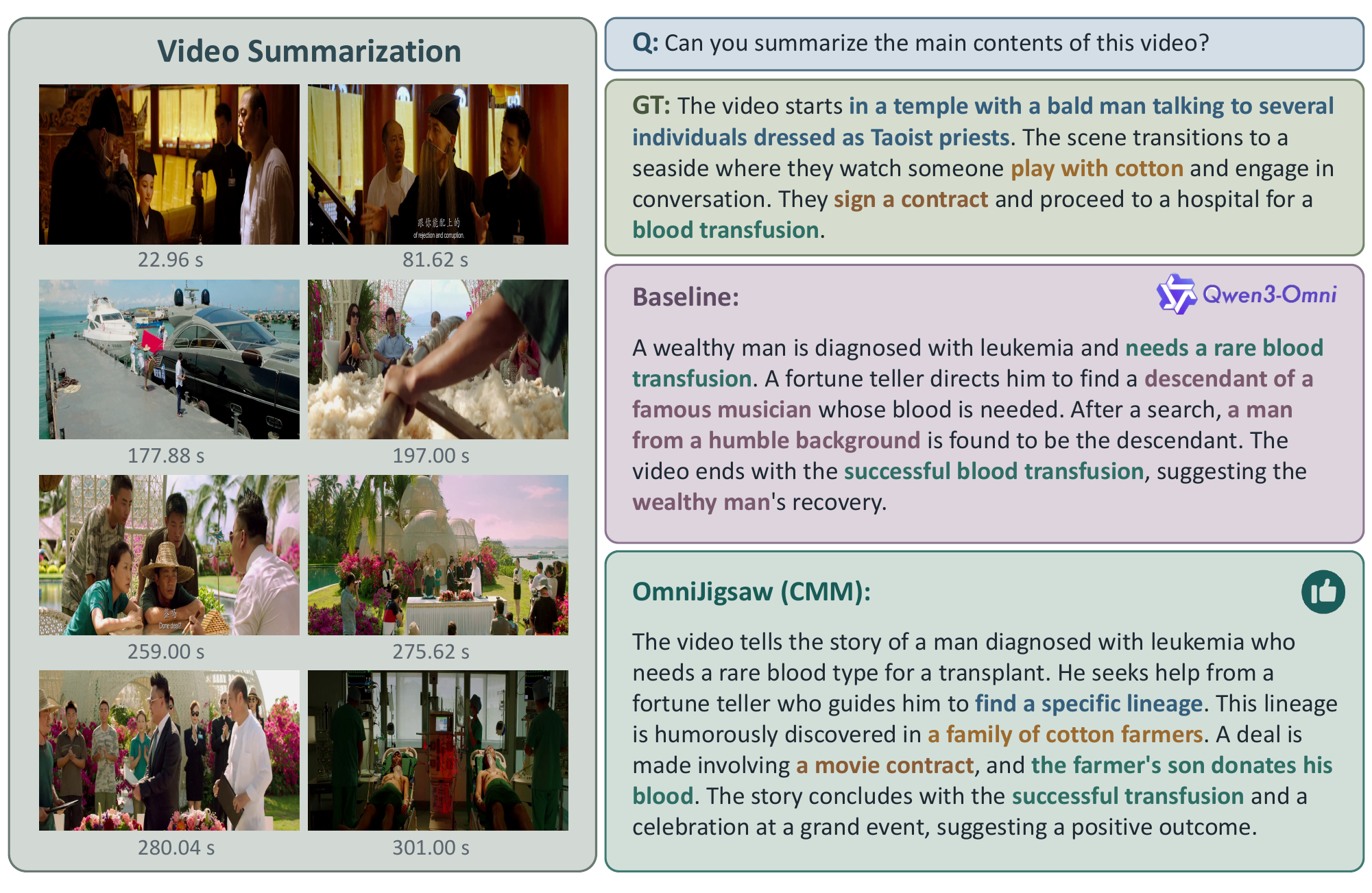}
    \caption{\textbf{Qualitative example of Video Summarization.} Comparison between the Qwen3-Omni-30B-A3B-Instruct~\cite{xu2025qwen3} baseline and its OmniJigsaw (CMM)-post-trained variant.}
    \label{fig:qual_video_summarization}
\end{figure*}

\subsection{Qualitative Examples}
\label{sec:supp_qual_examples}

Figures~\ref{fig:qual_subscene_captioning} and~\ref{fig:qual_video_summarization} show qualitative comparisons between the Qwen3-Omni-30B-A3B-Instruct~\cite{xu2025qwen3} baseline and its OmniJigsaw (CMM)-post-trained variant on two representative downstream tasks: Sub-Scene Captioning and Video Summarization. OmniJigsaw (CMM) consistently generates responses that are more faithfully grounded in the observed video content, while reducing unsupported speculation and improving long-horizon semantic coherence. As shown in Fig.~\ref{fig:qual_subscene_captioning}, the baseline drifts toward a post-incident description, focusing on actions such as ``knelt down beside her'' and ``examined it closely'', while failing to capture the decisive rescue chain required by the question. After OmniJigsaw (CMM) post-training, the model produces a more faithful interpretation of the event, explicitly identifying that the man ``grabbed the woman's arm'' and ``prevented her from falling'', which is substantially better aligned with the ground-truth explanation and the fine-grained scoring points. As illustrated in Fig.~\ref{fig:qual_video_summarization}, the baseline produces a more speculative narrative with unsupported details such as ``wealthy man'', ``descendant of a famous musician'', and ``humble background'', revealing weaker grounding in the observed sequence. After OmniJigsaw (CMM) post-training, the summary becomes considerably more faithful to the video storyline, capturing a much more coherent event chain involving ``a specific lineage'', ``a family of cotton farmers'', ``a contract'', and ``blood donation/transfusion''. These qualitative examples indicate that our modality-orchestrated reordering post-training strategy enhances both fine-grained action grounding and long-horizon narrative composition, while reducing unsupported speculation and strengthening the model's ability to associate temporally distributed clues into coherent semantic abstractions.

\section{Limitations and Future Work}
\label{sec:limitations_future_work}

While OmniJigsaw demonstrates consistent gains across video, audio, and omni-modal collaborative reasoning, several aspects need to be further explored. First, due to resource constraints and efficiency considerations, our study is conducted under a relatively conservative training setup on a single base model. Investigating its scalability and transferability across different model sizes, data scales, training settings, and omni-modal architectures is a valuable next step. Second, the current data curation pipeline is performed offline and therefore cannot explicitly adapt puzzle difficulty to the model’s evolving capabilities during post-training. Capability-aware or curriculum-based filtering that better matches different training stages merits further exploration. Third, our proxy task still relies on a relatively simple chronological reordering formulation with uniformly segmented clips. Richer puzzle designs such as variable clip lengths, overlapping clips, and hybrid spatio-temporal reordering deserve further study. Fourth, the current reward mainly emphasizes positional and adjacency correctness. Incorporating richer structure-aware or reasoning-aware supervision may provide more informative optimization signals. Finally, beyond temporal reordering, it is also worthwhile to explore a broader family of self-supervised omni-modal proxy tasks that intrinsically require the joint extraction and analysis of intertwined video and audio cues, offering a promising path toward more robust and capable omni-modal models through post-training on strong base models.

\begin{table}[t!]
    \centering
    \caption{\textbf{Sub-capability evaluation results on Video-MME~\cite{fu2025video}.} \gain{Green Bold} indicates performance surpassing the baseline.}
    \label{tab:video_mme_sub}
    \fontfamily{ptm}\selectfont
    \renewcommand{\arraystretch}{1.4}
    \setlength{\tabcolsep}{1.2pt} 
    \footnotesize
    \resizebox{\textwidth}{!}{
    \begin{tabular}{@{} l *{12}{c} @{}}
        \toprule
        \textbf{Method} & 

        \rotatebox{80}{\shortstack[l]{\textbf{Temporal}\\\textbf{Perception}}} & 
        \rotatebox{80}{\shortstack[l]{\textbf{Spatial}\\\textbf{Perception}}} & 
        \rotatebox{80}{\shortstack[l]{\textbf{Attribute}\\\textbf{Perception}}} & 
        \rotatebox{80}{\shortstack[l]{\textbf{Action}\\\textbf{Recognition}}} & 
        \rotatebox{80}{\shortstack[l]{\textbf{Object}\\\textbf{Recognition}}} & 
        \rotatebox{80}{\shortstack[l]{\textbf{OCR}\\\textbf{Problems}}} & 
        \rotatebox{80}{\shortstack[l]{\textbf{Counting}\\\textbf{Problem}}} & 
        \rotatebox{80}{\shortstack[l]{\textbf{Temporal}\\\textbf{Reasoning}}} & 
        \rotatebox{80}{\shortstack[l]{\textbf{Spatial}\\\textbf{Reasoning}}} & 
        \rotatebox{80}{\shortstack[l]{\textbf{Action}\\\textbf{Reasoning}}} & 
        \rotatebox{80}{\shortstack[l]{\textbf{Object}\\\textbf{Reasoning}}} & 
        \rotatebox{80}{\shortstack[l]{\textbf{Information}\\\textbf{Synopsis}}} \\
        \midrule
        \rowcolor{gray!15} Qwen3-Omni-30B & 80.0 & 70.4 & 77.0 & 69.0 & 76.6 & 74.1 & 46.3 & 57.1 & 78.6 & 57.2 & 65.9 & 80.5 \\
        OmniJigsaw (JMI) & 78.2 & \gain{74.1} & \gain{78.8} & \gain{71.2} & 74.6 & \gain{75.5} & \gain{47.4} & \gain{62.7} & 76.8 & \gain{58.6} & 65.9 & \gain{80.8} \\
        OmniJigsaw (SMS) & 80.0 & \gain{72.2} & \gain{80.2} & \gain{71.2} & 74.6 & 74.1 & \gain{48.1} & \gain{61.6} & 73.2 & \gain{59.3} & \gain{67.0} & 80.5 \\
        OmniJigsaw (CMM) & 76.4 & \gain{75.9} & \gain{78.8} & \gain{73.8} & 76.3 & \gain{77.0} & 44.4 & \gain{62.1} & 73.2 & \gain{61.8} & 65.9 & \gain{80.8} \\
        \bottomrule
    \end{tabular}
    }
\end{table}

\begin{table}[t!]
    \centering
    \caption{\textbf{Sub-capability evaluation results on MMAU-test-mini~\cite{sakshi2024mmau}.} \gain{Green Bold} indicates performance surpassing the baseline.}
    \label{tab:mmau_sub}
    \fontfamily{ptm}\selectfont
    \renewcommand{\arraystretch}{1.4}
    \setlength{\tabcolsep}{2.5pt}
    \footnotesize
    \begin{tabular}{@{} l ccc ccc @{}}
        \toprule
        & \multicolumn{3}{c}{\textbf{Task-wise}} & \multicolumn{3}{c}{\textbf{Difficulty-wise}} \\
        \cmidrule(lr){2-4} \cmidrule(lr){5-7}
        \textbf{Method} & \textbf{Sound} & \textbf{Music} & \textbf{Speech} & \textbf{Easy} & \textbf{Hard} & \textbf{Medium} \\
        \midrule
        \rowcolor{gray!15} Qwen3-Omni-30B & 76.88 & 72.75 & 73.57 & 70.09 & 70.76 & 77.78 \\
        OmniJigsaw (JMI) & \gain{77.18} & \gain{73.65} & 72.67 & \gain{70.54} & 68.22 & \gain{78.89} \\
        OmniJigsaw (SMS) & \gain{78.98} & \gain{73.65} & \gain{74.77} & \gain{70.98} & 68.64 & \gain{80.93} \\
        OmniJigsaw (CMM) & \gain{79.28} & \gain{75.15} & \gain{74.47} & \gain{73.66} & \gain{72.03} & \gain{79.26} \\
        \bottomrule
    \end{tabular}
\end{table}

\begin{table}[t!]
    \centering
    \caption{\textbf{Sub-capability evaluation results on OmniVideoBench~\cite{li2025omnivideobench}.} \gain{Green Bold} indicates performance surpassing the baseline.}
    \label{tab:omnivideobench_sub}
    \fontfamily{ptm}\selectfont
    \renewcommand{\arraystretch}{1.4}
    \setlength{\tabcolsep}{1.2pt}
    \footnotesize
    \resizebox{\textwidth}{!}{
    \begin{tabular}{@{} l *{13}{c} @{}}
        \toprule
        \textbf{Method} & 
        \rotatebox{80}{\shortstack[l]{\textbf{Attribute}\\\textbf{Comparison}}} & 
        \rotatebox{80}{\shortstack[l]{\textbf{Bkg. \& Music}\\\textbf{Understanding}}} & 
        \rotatebox{80}{\shortstack[l]{\textbf{Causal}\\\textbf{Reasoning}}} & 
        \rotatebox{80}{\textbf{Counting}} & 
        \rotatebox{80}{\shortstack[l]{\textbf{Ego}\\\textbf{Reasoning}}} & 
        \rotatebox{80}{\shortstack[l]{\textbf{Fine-grained}\\\textbf{Perception}}} & 
        \rotatebox{80}{\shortstack[l]{\textbf{Hypothetical}\\\textbf{Reasoning}}} & 
        \rotatebox{80}{\shortstack[l]{\textbf{Reference}\\\textbf{Reasoning}}} & 
        \rotatebox{80}{\shortstack[l]{\textbf{Relationship}\\\textbf{Reasoning}}} & 
        \rotatebox{80}{\shortstack[l]{\textbf{Sentiment}\\\textbf{Analysis}}} & 
        \rotatebox{80}{\shortstack[l]{\textbf{Spatial}\\\textbf{Understanding}}} & 
        \rotatebox{80}{\textbf{Summarization}} & 
        \rotatebox{80}{\shortstack[l]{\textbf{Temporal}\\\textbf{Understanding}}} \\
        \midrule
        \rowcolor{gray!15} Qwen3-Omni-30B & 28.57 & 34.04 & 47.14 & 25.16 & 45.07 & 38.46 & 58.33 & 41.84 & 70.83 & 38.27 & 22.58 & 63.27 & 33.58 \\
        OmniJigsaw (JMI) & \gain{61.90} & 31.91 & \gain{49.29} & 21.29 & \gain{54.93} & \gain{47.25} & 45.83 & \gain{45.92} & 66.67 & 38.27 & 22.58 & 55.10 & 32.85 \\
        OmniJigsaw (SMS) & \gain{38.10} & 27.66 & \gain{50.00} & \gain{26.45} & \gain{47.89} & \gain{41.76} & 41.67 & 37.76 & \gain{79.17} & \gain{44.44} & \gain{30.65} & 61.22 & \gain{34.31} \\
        OmniJigsaw (CMM) & \gain{38.10} & 29.79 & 46.43 & 24.52 & 43.66 & \gain{41.76} & 58.33 & \gain{44.90} & 70.83 & \gain{41.98} & \gain{32.26} & 55.10 & \gain{40.15} \\
        \bottomrule
    \end{tabular}
    }
\end{table}

\section{Prompts}
\label{sec:prompts}

\subsection{Semantic Screening Prompt}
\label{sec:semantic_screening_prompt}

\begin{promptbox}{Prompt for Semantic Screening}
    You are an expert video analyst tasked with determining if a video is suitable for a ``Video Jigsaw'' (temporal reordering) task.

    \vspace{1\baselineskip}
    
    \textbf{Task Definition:} 
    
    A video is suitable (YES) if it contains a clear, \textbf{directionally unambiguous} chronological progression with \textbf{distinct visual states}. If this video were cut into 3-6 segments and shuffled, a human should be able to restore the original order by identifying specific visual cues and logical flow.

    \vspace{1\baselineskip}
    
    \textbf{Evaluation Criteria:}
    
    1. \textbf{Causal Progression:} Does action A lead to result B? (\eg, ingredients being mixed then put into an oven).
    
    2. \textbf{Visual State Changes:} Is there a clear evolution of an object or environment? (\eg, a house being built, a drawing being completed, a plant growing).
    
    3. \textbf{Logical Narrative:} Is there a sequential flow of events? (\eg, a person walking into a building, then sitting at a desk).
    
    4. \textbf{Temporal Markers:} Are there environmental changes like lighting (day to night) or physical wear/consumption?

    \vspace{1\baselineskip}
    
    \textbf{Exclusion Criteria (Answer NO if):}
    
    1. \textbf{Repetitive Loops:} The person is doing the same action over and over (\eg, a gym workout loop, a person just nodding/talking without changing position).
    
    2. \textbf{Static/Low-Dynamic Content:} Talking heads with static backgrounds, or videos with minimal movement/change.
    
    3. \textbf{Parallel/Unrelated Clips:} A compilation of unrelated highlights where order doesn't matter.
    
    4. \textbf{Visual Ambiguity:} Different parts of the video look too similar (\eg, a person walking in a desert with no landmarks), making reordering a guess.

    \vspace{1\baselineskip}
    
    \textbf{Output Format:}
    
    You MUST follow this format:
    
    <think>
    
    1. Briefly describe the key actions and state changes observed.
    
    2. Evaluate against the Evaluation and Exclusion criteria.
    
    3. Conclude whether the video is suitable for the ``Video Jigsaw'' task.
    
    </think>
    
    <answer>YES or NO</answer>
\end{promptbox}

\clearpage

\subsection{Training Prompts}
\label{sec:training_prompts}
In practice, since Qwen3-Omni-30B-A3B-Instruct~\cite{xu2025qwen3} does not reliably output \texttt{<think>} during generation, likely due to the tokenizer behavior or pretraining treatment associated with this tag, we use the semantically equivalent \texttt{<thinking>} tag in the training and evaluation prompts to improve format adherence.
\begin{promptbox}{Prompt for OmniJigsaw (JMI)}
    You are given 6 \textbf{shuffled} video clips (with audio) that were created by slicing one original video into 6 equal-length temporal segments.

    \vspace{1\baselineskip}

    Here are the clips, each tagged with an index reflecting the current (shuffled) order in which they are shown. Please pay attention to both the visual actions and the specific content of the audio.

    \vspace{1\baselineskip}
    
    Clip 1: <video>
    
    Clip 2: <video>
    
    Clip 3: <video>
    
    Clip 4: <video>
    
    Clip 5: <video>
    
    Clip 6: <video>

    \vspace{1\baselineskip}
    
    Task:
    
    1. \textbf{For each clip}, describe its specific visual events and audio features (\eg, distinct speech phrases or sound patterns).
    
    2. Analyze the temporal logic by comparing these features to determine how the story or action progresses from one segment to another.
    
    3. Based on this content-driven analysis, reassemble the original video and output the indices in correct order, separated by commas.
    
    \vspace{1\baselineskip}
    
    Answer format example:
    
    2, 3, 1, 4, 6, 5

    \vspace{1\baselineskip}
    
    You \textbf{FIRST} think about the reasoning process as an internal monologue and \textbf{THEN} provide the final answer. \textbf{Ensure your reasoning is strictly based on audio-visual facts and follows a logical progression, avoiding unrelated digressions.} The reasoning process MUST BE enclosed within <thinking> </thinking> tags. The final answer MUST BE enclosed within <answer> </answer> tags.
\end{promptbox}

\begin{promptbox}{Prompt for OmniJigsaw (SMS) (Dominance Analyzer)}
    \#\#\# Role
    
    You are a Multimodal Content Analyst. Your goal is to evaluate the \textbf{visual stream} and \textbf{audio track} of the provided video to determine the optimal modality for a ``Temporal Jigsaw Puzzle''.

    \vspace{1\baselineskip}
    
    \#\#\# Task
    
    You must decide whether the \textbf{Visual stream} or the \textbf{Audio stream} contains stronger, more deterministic clues for reconstructing the chronological order if this video were sliced into shuffled clips.

    \vspace{1\baselineskip}
    
    \#\#\# Analysis Guidelines
    
    \textbf{1. Evaluate Visual Suitability (For Video Jigsaw):}
    
    *   \textbf{Look for:} Clear physical actions (\eg, pouring water), scene transitions, camera movement, or object state changes (\eg, assembling a puzzle).
    
    *   \textbf{Penalty:} Give low priority to ``V'' if the video is static (\eg, a person sitting still with minimal movement), repetitive (loops), mostly black/blurry, or shows minimal change from beginning to end.
    
    \textbf{2. Evaluate Audio Suitability (For Audio Jigsaw):}
    
    *   \textbf{Look for:} Continuous Speech (\eg, narrative/dialogue with logical flow), Musical Progression (\eg, verse -> chorus), or Sequential Sound Events (\eg, footsteps -> door open -> slam).
    
    *   \textbf{Penalty:} Give low priority to ``A'' if the audio is constant background noise, a repetitive music loop without progression, or silence.
    
    \textbf{3. Comparison \& Decision:}
    
    *   \textbf{Select ``V''} if the visual evolution provides a stricter, unambiguous timeline (\eg, a silent movie with clear plot actions).
    
    *   \textbf{Select ``A''} if the auditory narrative provides the primary logical thread (\eg, a podcast, a speech, or a static shot of a narrator).
    
    *   \textit{Tie-Breaker:} If both are good, choose the one that requires less ambiguous guessing for solving the ``Temporal Jigsaw Puzzle''.

    \vspace{1\baselineskip}
    
    \#\#\# Input Data
    
    <video>

    \vspace{1\baselineskip}
    
    \#\#\# Output Format
    
    Output \textbf{only} the single character decision inside valid tags.
    
    <answer>V</answer>
    
    or
    
    <answer>A</answer>
    
    You \textbf{FIRST} think about the comparison process as an internal monologue \textbf{(Analyze Visual Suitability -> Analyze Audio Suitability -> Final Decision)} and \textbf{THEN} provide the final answer. The thinking process MUST BE enclosed within <thinking> </thinking> tags. The final answer MUST BE enclosed within <answer> </answer> tags.
\end{promptbox}

\begin{promptbox}{Prompt for VideoJigsaw \& OmniJigsaw (SMS) (Video Jigsaw Rollout)}
    You are given 6 \textbf{shuffled} video clips that were created by slicing one original video into 6 equal-length temporal segments.

    \vspace{1\baselineskip}
    
    Here are the clips, each tagged with an index reflecting the current (shuffled) order in which they are shown. Please pay attention to both the visual actions and specific content.
    
    \vspace{1\baselineskip}
    
    Clip 1: <video>
    
    Clip 2: <video>
    
    Clip 3: <video>
    
    Clip 4: <video>
    
    Clip 5: <video>
    
    Clip 6: <video>

    \vspace{1\baselineskip}
    
    Task:
    
    1. \textbf{For each clip}, describe its specific visual events and features (\eg, distinct event phases or evolving visual states).
    
    2. Analyze the temporal logic by comparing these features to determine how the story or action progresses from one segment to another.
    
    3. Based on this content-driven analysis, reassemble the original video and output the indices in correct order, separated by commas.
    
    \vspace{1\baselineskip}
    
    Answer format example:
    
    2, 3, 1, 4, 6, 5
    
    \vspace{1\baselineskip}
    
    You \textbf{FIRST} think about the reasoning process as an internal monologue and \textbf{THEN} provide the final answer. \textbf{Ensure your reasoning is strictly based on visual facts and follows a logical progression, avoiding unrelated digressions.} The reasoning process MUST BE enclosed within <thinking> </thinking> tags. The final answer MUST BE enclosed within <answer> </answer> tags.
\end{promptbox}

\begin{promptbox}{Prompt for AudioJigsaw \& OmniJigsaw (SMS) (Audio Jigsaw Rollout)}
    You are given 6 \textbf{shuffled} audio clips that were created by slicing one original audio into 6 equal-length temporal segments.

    \vspace{1\baselineskip}
    
    Here are the clips, each tagged with an index reflecting the current (shuffled) order in which they are shown. Please pay attention to both the specific auditory events and content.
    
    \vspace{1\baselineskip}
    
    Clip 1: <audio>
    
    Clip 2: <audio>
    
    Clip 3: <audio>
    
    Clip 4: <audio>
    
    Clip 5: <audio>
    
    Clip 6: <audio>

    \vspace{1\baselineskip}
    
    Task:
    
    1. \textbf{For each clip}, \textbf{briefly} describe its specific auditory events and features (\eg, distinct speech phrases or sound patterns).
    
    2. Analyze the temporal logic by comparing these features to determine how the speech context, musical progression, or sound events evolve from one segment to another.
    
    3. Based on this content-driven analysis, reassemble the original audio and output the indices in correct order, separated by commas.

    \vspace{1\baselineskip}
    
    Answer format example:
    
    2, 3, 1, 4, 6, 5

    \vspace{1\baselineskip}
    
    You \textbf{FIRST} think about the reasoning process as an internal monologue and \textbf{THEN} provide the final answer.
    
    \textbf{Thinking Process Constraints:}
    
    1. \textbf{Be Concise:} Your entire thinking process MUST NOT exceed \textbf{800 words}.
    
    2. \textbf{Strictly Auditory:} Ensure your reasoning is strictly based on auditory facts and follows a logical progression, avoiding unrelated digressions.
    
    The reasoning process MUST BE enclosed within <thinking> </thinking> tags. The final answer MUST BE enclosed within <answer> </answer> tags.
\end{promptbox}

\begin{promptbox}{Prompt for OmniJigsaw (CMM) (Modality Selector)}
    \#\#\# Role
    
    You are a Multimodal Jigsaw Puzzle Expert. Your goal is to curate the modalities for a high-quality ``OmniJigsaw'' puzzle based on 6 sequential video clips.
    
    \vspace{1\baselineskip}
    
    \#\#\# What is OmniJigsaw?
    
    OmniJigsaw challenges a solver to reconstruct the chronological timeline of shuffled clips using a curated mix of Visual (V), Audio (A), and Cross-modal (VA) clues.

    \vspace{1\baselineskip}
    
    \#\#\# Input Data
    
    You are given 6 \textbf{ordered} video clips (each contains both video and audio). These clips were extracted from one original video in chronological order and have equal length.
    
    Clip 1: <video> Clip 2: <video> Clip 3: <video> Clip 4: <video> Clip 5: <video> Clip 6: <video> 

    \vspace{1\baselineskip}
    
    \#\#\# Task
    
    Assign the most appropriate modality (\textbf{``V''}, \textbf{``A''}, or \textbf{``VA''}) to \textbf{each} clip based on its information richness and its role in the temporal sequence.
    
    \textbf{Analysis Steps:}
    
    1.  \textbf{Content Evaluation:} For each clip, assess the richness and distinctiveness of its Visual stream vs. its Audio stream.
    
    2.  \textbf{Modality Decision Logic:}
    
        - \textbf{Select ``V'':} If the \textbf{Visual} evidence is distinct (\eg, scene change, physical motion) while the audio is generic, ambient, or redundant.
        
        - \textbf{Select ``A'':} If the \textbf{Audio} evidence is distinct (\eg, continuous sentence, distinct sound cue) while the visual is static, blurry, or less informative.
        
        - \textbf{Select ``VA'':} If \textbf{BOTH} video and audio provide rich, critical clues for placing this clip in the timeline (\eg, a specific character action synchronized with a plot-critical dialogue). Use VA when the clip serves as a strong ``anchor'' that requires full context.
        
    3.  \textbf{Sequence Verification:}
    
        - Review your selections. Do they allow the solver to deduce the order through natural content progression? Ensure the sequence isn't rendered unsolvable by over-hiding critical cues.

    \vspace{1\baselineskip}
    
    \#\#\# Constraints
    
    - \textbf{Avoid} outputting all ``V'', all ``A'', or all ``VA''.
    
    - Strive for a natural mix. While VA is valuable, distinct V and A clips are also essential for testing specific modality understanding.

    \vspace{1\baselineskip}
    
    \#\#\# Output Format
    
    You MUST provide the final result in valid JSON format inside <answer>...</answer> tags following this structure:
    
    \texttt{\{"modalities": ["V", "A", "VA", "V", "A", "V"]\}}
    
    You \textbf{FIRST} think about the \textbf{Content Richness and Modality Strategy} as an internal monologue (Content Evaluation -> Modality Selection -> Sequence Verification) and \textbf{THEN} provide the final design. The thinking process MUST BE enclosed within <thinking> </thinking> tags. The final design MUST BE enclosed within <answer> </answer> tags.
\end{promptbox}

\begin{promptbox}{Prompt for OmniJigsaw (CMM) (Jigsaw Rollout)}
    You are given 6 \textbf{shuffled} clips that were created by slicing one original video into 6 equal-length temporal segments.

    \vspace{1\baselineskip}
    
    \textbf{Important Note on Modalities:} To test your cross-modal understanding, some clips may have missing audio (the audio is silent) or missing video (the video frames are entirely black). Use the non-missing and informative modalities in each clip to identify its content.

    \vspace{1\baselineskip}
    
    Here are the clips, each tagged with an index reflecting the current (shuffled) order in which they are shown. Please pay attention to both the visual facts and auditory clues.
    
    \vspace{1\baselineskip}
    
    Clip 1: <video>
    
    Clip 2: <video>
    
    Clip 3: <video>
    
    Clip 4: <video>
    
    Clip 5: <video>
    
    Clip 6: <video>

    \vspace{1\baselineskip}
    
    Task:
    
    1. \textbf{For each clip}, describe its specific visual events and audio features (\eg, distinct speech phrases or sound patterns). If a modality is missing, focus on the available one. \textbf{For clip connections}, identify subtle transition clues (\eg, action/state continuity, unfinished sentences, background sounds) and analyze the causal, logical, or emotional progression.
    
    2. Analyze the temporal logic by integrating individual clip content with inter-clip relationships, explicitly chaining together both specific continuity details and thematic progression.
    
    3. Based on this evidence-driven analysis, reassemble the original video and output the indices in correct order, separated by commas.

    \vspace{1\baselineskip}
    
    Answer format example:
    
    2, 3, 1, 4, 6, 5

    \vspace{1\baselineskip}
    
    You \textbf{FIRST} think about the reasoning process as an internal monologue and \textbf{THEN} provide the final answer. \textbf{Ensure your reasoning is strictly based on clip content features, inter-clip relationships, and specific connection evidence, avoiding unrelated digressions.} The reasoning process MUST BE enclosed within <thinking> </thinking> tags. The final answer MUST BE enclosed within <answer> </answer> tags.
\end{promptbox}

\subsection{Evaluation Prompts}
\label{sec:evaluation_prompts}

\begin{promptbox}{Prompt Template for Video Reasoning}
    Please answer the following question based on the complete video content.

    \vspace{1\baselineskip}

    <task>
    
    \textbf{Question:} \{question\}
    
    \textbf{Candidates:} \{candidates\}
    
    </task>

    \vspace{1\baselineskip}
    
    You \textbf{FIRST} think about the reasoning process as an internal monologue and \textbf{THEN} provide the final answer. The reasoning process MUST BE enclosed within <thinking> </thinking> tags. The final answer (just the option letter) MUST BE enclosed within <answer> </answer> tags.
\end{promptbox}

\begin{promptbox}{Prompt Template for Audio Reasoning}
    Please answer the following question based on the complete audio content.

    \vspace{1\baselineskip}

    <task>
    
    \textbf{Question:} \{question\}
    
    \textbf{Candidates:} \{candidates\}
    
    </task>

    \vspace{1\baselineskip}
    
    You \textbf{FIRST} think about the reasoning process as an internal monologue and \textbf{THEN} provide the final answer. The reasoning process MUST BE enclosed within <thinking> </thinking> tags. The final answer (just the option letter) MUST BE enclosed within <answer> </answer> tags.
\end{promptbox}

\begin{promptbox}{Prompt Template for Omni-Modal Collaborative Reasoning}
    Please answer the following question based on the complete video content (visual + audio).

    \vspace{1\baselineskip}
    
    <task>
    
    \textbf{Question:} \{question\}
    
    \textbf{Candidates:} \{candidates\}
    
    </task>

    \vspace{1\baselineskip}
    
    You \textbf{FIRST} think about the reasoning process as an internal monologue and \textbf{THEN} provide the final answer. The reasoning process MUST BE enclosed within <thinking> </thinking> tags. The final answer (just the option letter) MUST BE enclosed within <answer> </answer> tags.
\end{promptbox}
\end{document}